\documentclass{article}

\PassOptionsToPackage{numbers, compress}{natbib}

\usepackage[main, final]{neurips_2026}

\usepackage[utf8]{inputenc} 
\usepackage[T1]{fontenc}    
\usepackage{hyperref}       
\usepackage{url}            
\usepackage{booktabs}       
\usepackage{amsfonts}       
\usepackage{nicefrac}       
\usepackage{microtype}      
\usepackage{xcolor}         

\usepackage[table]{xcolor}
\usepackage[disable,textsize=tiny]{todonotes}
\usepackage[textsize=tiny]{todonotes}

\usepackage{mathtools}
\usepackage{amsthm}
\usepackage{multirow}
\usepackage{blindtext}
\usepackage{graphicx}
\usepackage{makecell}
\usepackage{enumitem}
\usepackage{algorithm}
\usepackage{algorithmic}
\usepackage{tikz}
\usepackage{subcaption}
\usepackage{csquotes}
\usepackage{wrapfig}
\newcommand{\minisection}[1]{\vspace{5pt}\noindent\textbf{#1.}}

\newcommand{\method}{\textsc{Kairos}\xspace}

\usepackage{bbding}
\usepackage[figuresright]{rotating}
\usepackage{array}
\usepackage{pifont}

\theoremstyle{plain}
\newtheorem{theorem}{Theorem}[section]

\theoremstyle{definition}
\newtheorem{definition}[theorem]{Definition}

\theoremstyle{remark}

\usepackage{wrapfig}

\title{\method: Toward Adaptive and Parameter-Efficient Time Series Foundation Models}

\author{
\textbf{Kun Feng$^{1}$\thanks{
Equal contribution.
} ,
Shaocheng Lan$^{1}$\footnotemark[1] ,
Yuchen Fang$^{2}$\footnotemark[1] ,
Wenchao He$^{1}$}
\\
\textbf{Sihan Lu$^{1}$, Shuqi Gu$^{1}$, Xingyu Lu$^{2}$, Lintao Ma$^{2}$,}
\textbf{Kan Ren$^{1}$\thanks{Corresponding author: \texttt{renkan@shanghaitech.edu.cn}}} \\
$^{1}$ShanghaiTech University \quad $^{2}$Ant Group
\\
\texttt{\{fengkun2025,lanshch2024,renkan\}@shanghaitech.edu.cn}
\\
\texttt{yuchen.fyc@antgroup.com}
}

\begin{document}

\maketitle

\begin{abstract}
Inherent temporal heterogeneity, such as varying sampling densities and periodic structures, has posed substantial challenges in zero-shot generalization for Time Series Foundation Models (TSFMs).
Existing TSFMs predominantly rely on massive parameterization to absorb such heterogeneity, as their static tokenization and positional encoding schemes entangle diverse temporal patterns into a fixed representation space, encouraging memorization rather than adaptation.
To address this limitation, we propose \method, a flexible and parameter-efficient TSFM dedicated to forecasting tasks, which decouples temporal heterogeneity from model capacity through a novel tokenization perspective. 
\method introduces a dynamic patching tokenizer and a mixture-of-size encoding that adapt observational granularity to local information density, enabling fine-grained temporal abstraction without increasing model width or depth. 
In addition, we design a multi-granularity positional embedding based on dynamic rotary encodings, which conditions on instance-level spectral features and temporal structure induced by dynamic patching tokenization, 
allowing robust modeling of diverse temporal dependencies.
Trained on a novel Predictability-Stratified Time-Series (PreSTS) corpus, 
\method achieves superior zero-shot performance with substantially fewer parameters on two mainstream benchmarks, GIFT-Eval and Time-Series-Library.
The project page is at \url{https://foundation-model-research.github.io/Kairos}.
\end{abstract}

\begin{figure*}[h]
\centering
\includegraphics[width=\textwidth]{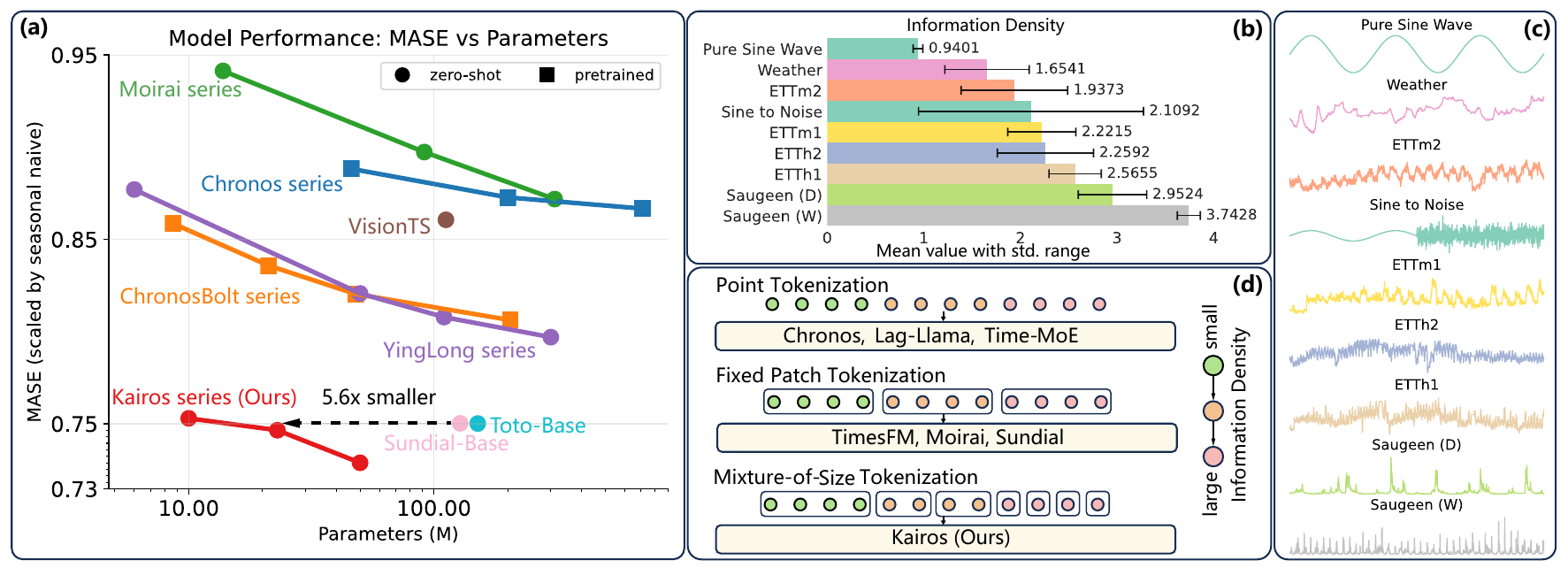}
\caption{\textbf{(a)}
Comparison on the GIFT-Eval benchmark~\citep{aksu2024gifteval} demonstrates that our \method achieves superior zero-shot forecasting performance (lower normalized MASE) while requiring significantly fewer parameters than existing TSFMs. 
\textbf{(b) (c)} Significant variation exists in information density (i.e., signal complexity) across and within different time series datasets.
\textbf{(d)} 
Existing TSFMs primarily use tokenization methods like point-wise or fixed-size patching, while our \method uses Mixture-of-Size Tokenization to dynamically adapt to information density, enabling the learning of generalizable rules rather than memorization.
}
\label{fig:intro}
\end{figure*}

\section{Introduction}
\label{sec:intro}

Time series forecasting is a core component of many real-world applications~\citep{box2015time,wang2024tssurvey}.
Although time series data are ubiquitous, many individual datasets suffer from scarcity, making it difficult to train models independently, particularly for deep neural networks that require large amounts of data.
This challenge has long motivated the development of few-shot and zero-shot forecasting methods~\citep{Oreshkin2020:N-BEATS}.
More recently, inspired by the success of large-scale foundation models~\citep{achiam2023gpt,kirillov2023segany}, researchers have turned to building 
Time Series Foundation Models (TSFMs) for forecasting.
These models leverage vast and diverse time series corpora to acquire strong generalization capabilities, showing promising results across various downstream tasks~\citep{ansari2024chronos,woo2024moirai,shi2024timemoe}.

Despite this progress, a key limitation persists: current TSFMs for forecasting rely on non-adaptive processing pipelines that fail to reflect the dynamic and heterogeneous nature of time series, severely compromising their parameter efficiency. 
As shown in Figure \ref{fig:intro}(b)(c), the statistical information density (details in Appendix~\ref{sec:cal_information_density}) varies not only across datasets but also across segments within the same series.
By applying rigid architectures (exemplified in Figure \ref{fig:intro}(d)) to this inherently heterogeneous data, models are forced to use massive parameters to memorize specific patterns rather than acquiring adaptive, generalizable rules. 
Unfortunately, this strategy diminishes the overall utility of the model's capacity, rendering it less effective for real-world forecasting applications.

This deficiency manifests primarily in two fundamental aspects of model design:
(i) \emph{Representation learning}:
TSFMs have yet to adopt dynamic \emph{tokenization strategies} analogous to those in Large Language Models (LLMs). 
While LLMs utilize subword tokenization~\citep{gage1994new} to efficiently compress semantic units of varying lengths, TSFMs predominantly persist with static patching, whether through point-wise tokenization~\citep{ansari2024chronos,shi2024timemoe} or fixed-size patches~\citep{das2024decoder}.
By coupling token generation to fixed time intervals rather than information content, these approaches compress diverse temporal variations into a rigid representation space.
Consequently, the model wastes computational capacity on low-information segments while struggling to capture high-frequency details within the same fixed budget, leading to significant parameter inefficiency.
(ii) \emph{Temporal dependency modeling}:
Conventional \emph{positional encodings} in TSFMs impose a unified temporal scale across sequences, neglecting differences in periodicity, trend and seasonality.
As a result, existing TSFMs struggle to adapt to heterogeneous domains, for example, power consumption data measured hourly versus retail sales data measured daily, each exhibiting distinct temporal dynamics, as we validate in Section~\ref{sec:ablation_study}.
Consequently, the model requires a deeper, more complex architecture to compensate for this misalignment, further exacerbating parameter inefficiency and hindering practical utility.

To address these limitations, we propose \method, a parameter-efficient model comprising a Mixture-of-Size Encoder, a Heterogeneity-Aware Transformer, and a Multi-Patch Decoder. 
\method 
addresses time series heterogeneity through two primary mechanisms:
(i) For intra-patch representation learning, the Mixture-of-Size Encoder adaptively models time series at multiple granularities based on local characteristics.
Inspired by null experts in Mixture-of-Experts~\citep{zeng2024adamoe,jinmoe++}, we introduce computation-free null experts to dynamically adjust the number of active granularities.
This approach moves beyond static tokenization strategies, enabling more expressive modeling of nuanced, time-varying dynamics.
(ii) For inter-patch dependency modeling, we enhance the Transformer with Dynamic Rotary Position Embedding (DRoPE).
Unlike standard RoPE~\citep{su2024roformer}, which represents positional information with a fixed temporal scale, DRoPE modulates temporal scales using spectral features and granularity to generate instance-specific positional encodings.
Flexible intra-patch and inter-patch modeling enable \method to effectively characterize heterogeneous time series with a substantially reduced parameter count (see Figure~\ref{fig:intro}(a)).
Additionally, the Multi-Patch Decoder uses learnable tokens to predict future patches in parallel, which mitigates cumulative errors in autoregressive generation and offers greater flexibility for variable-length prediction horizons.

To complement our architecture with high-quality supervision, we construct the Predictability-Stratified Time Series (PreSTS) corpus, a large-scale and diverse pretraining dataset curated through a targeted sampling strategy. 
By prioritizing more predictable sequences while maintaining broad coverage, PreSTS provides high-quality supervision that supports efficient model scaling.
As demonstrated by the results on the GIFT-Eval benchmark~\citep{aksu2024gifteval} in Figure~\ref{fig:intro}(a), these combined innovations allow our \method to achieve superior performance while using fewer parameters. 
This strong performance is consistently observed on Time-Series-Library (TSLib)~\citep{wang2024tssurvey} benchmarks as well.

In summary, our main contributions are as follows: 
(i) We present \method, a parameter-efficient time series foundation model that introduces a novel architectural framework to handle the dynamic and heterogeneous nature of time series data.
(ii) We curate the PreSTS corpus, a large-scale pre-training dataset of over 300 billion time points, featuring a predictability-based sampling strategy to provide high-quality and diverse supervision.
(iii) Extensive empirical evaluations on the GIFT-Eval and Time-Series-Library benchmarks confirm the efficacy of \method, highlighting its superior zero-shot forecasting performance.

\section{Related Work}
\label{sec:related_work}
\textbf{Time Series Foundation Models (TSFMs) for Forecasting.} 
Inspired by the strong generalization of Large Language Models (LLMs), researchers have devoted increasing attention to TSFMs.
Existing approaches typically involve adapting LLMs to time series~\citep{gruver2023large}, training scalable Transformers on large corpora~\citep{ansari2024chronos,das2024decoder,woo2024moirai,cohen2025time,liu2025sundial}, or exploring alternative architectures, including non-Transformer models~\citep{auer2025tirex,graf2025flowstate} and lightweight Multi-Layer Perceptrons~\citep{ekambaram2024tiny}.
However, these models often apply fixed tokenization strategies and position embedding to time series from diverse domains with distinct characteristics, limiting their cross-domain generalization. 
\method, in contrast, employs a dynamic patching tokenizer and instance-adaptive, granularity-aware position embedding to enhance its generalizability for diverse time series data.

\textbf{Time Series Tokenization.} 
Transformer-based models require converting raw data into discrete or continuous tokens~\citep{sennrich-etal-2016-neural, agarwal2025cosmos}.
Current strategies generally fall into two static paradigms:
(i) \textit{Fixed-size}:
The dominant paradigm relies on a uniform patch size across the sequence. 
This ranges from fine-grained point-wise tokenization~\citep{ansari2024chronos,rasul2023lag,shi2024timemoe} to coarser uni-size patching~\citep{nie2023time}, which has been adopted by recent foundation models~\citep{woo2024moirai,liu2025sundial}.
While the latter improves upon the computational inefficiency of point-wise methods, both remain rigid.
(ii) \textit{Multi-size}:
To capture broader temporal dynamics, models like Pathformer~\citep{chen2024pathformer} and TTM~\citep{ekambaram2024tiny} incorporate multiple predefined patch sizes. 
However, these approaches apply selected granularities uniformly across the entire sequence based on global features rather than local information.
This rigidity forces the model to allocate equal computational cost to segments of varying complexity. 
This imbalance leads to suboptimal parameter efficiency and limited generalization. 
In contrast, \method employs a mixture-of-size strategy.
By allocating fewer tokens to simple patterns and more to complex regions, \method enhances parameter efficiency while maintaining robust generalization.

\textbf{Position Embedding.} 
Due to the position-unaware characteristic of self-attention mechanism, Transformers~\cite{vaswani2017attention} rely on position embeddings (PEs) to model temporal information.
However, most TSFMs adopt existing designs for PE in Natural Language Processing (NLP) domains~\citep{das2024decoder, shi2024timemoe}, which typically emphasize long-term decay through methods like sinusoidal calculations or by suppressing long-range positional information \citep{vaswani2017attention,su2024roformer}.
By imposing a uniform temporal scale, these PEs struggle to model the heterogeneous temporal relationships of time series. 
Existing adaptive PE methods in NLP~\citep{zheng2024dape,lin2024mixture} are also insufficient, as they target context extrapolation rather than the complex temporal structures of time series.
Therefore, ElasTST~\citep{zhang2024elastst} proposed a tunable RoPE to better adapt to time series data. 
However, its adaptation uses per-dataset training rather than dynamic adjustment. 
To address this limitation, we propose to dynamically modulate PE tailored to the intrinsic characteristics of each input time series, effectively decoupling temporal heterogeneity from model capacity, therefore enhancing the parameter efficiency and overall performance of TSFMs.
\begin{figure*}[h]
\centering
\includegraphics[width=\textwidth]{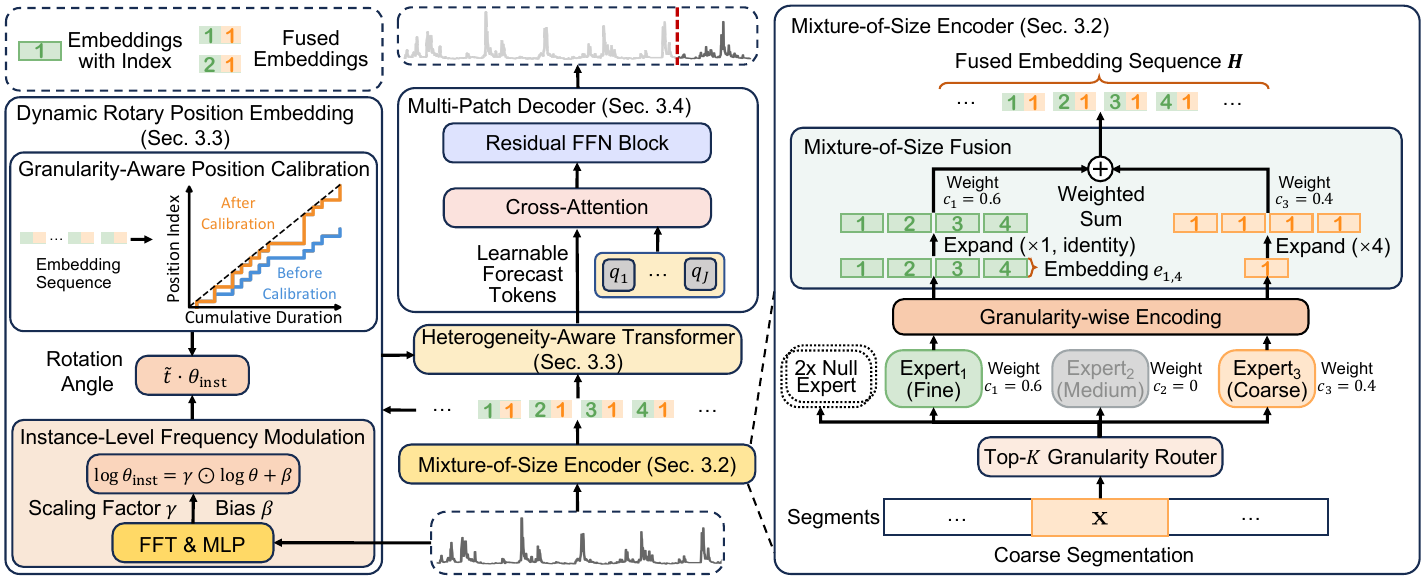}
\caption{
The architecture of \method, which includes 
(i) The Mixture-of-Size Encoder (right) adaptively tokenizes the input series by routing segments to optimal granularity experts and fusing their embeddings into a unified sequence $\boldsymbol{H}$.
Embedding subscripts identify the expert and patch within a segment (Eq.~\ref{eq:granularity_embedding}).
(ii) The Heterogeneity-Aware Transformer (middle) processes these tokens using Dynamic Rotary Position Embedding (DRoPE) (left), which modulates frequencies based on instance-level spectral features and calibrates positions to account for varying patch sizes. 
(iii) The Multi-Patch Decoder (top-middle) employs learnable forecast tokens to predict multiple future patches in parallel via cross-attention.
}
\label{fig:kairos}
\end{figure*}
\section{Methodology}
\label{sec:method}
\subsection{Overall Framework}
We formulate our task as learning a time series forecasting function $f_\phi: \mathbb{R}^T \to \mathbb{R}^H$ parameterized by $\phi$.
Given historical time series $\mathbf{X} \in \mathbb{R}^T$, time series forecasting is to predict the future horizon $\mathbf{Y} \in \mathbb{R}^H$.
The learning objective is to minimize the discrepancy between the ground truth $\mathbf{Y}$ and the prediction $\hat{\mathbf{Y}} = f_\phi(\mathbf{X})$.
We detail it in Appendix~\ref{app:loss_function}.

The architecture of \method, shown in Figure~\ref{fig:kairos}, consists of three components.
First, the input time series is tokenized by the Mixture-of-Size Encoder to extract multi-granularity local representations via dynamic granularity selection.
These embeddings are then processed by a Heterogeneity-Aware Transformer equipped with Dynamic Rotary Position Embedding (DRoPE), a granularity-aware positional encoding that performs instance-wise modulation and adapts to dynamic patching tokenization.
Finally, the Multi-Patch Decoder generates forecasts by predicting multiple future patches in parallel, mitigating cumulative errors.
The details of these components are presented in the following sections.

\subsection{Mixture-of-Size Encoder}
\label{sec:mos-encoder}

Prior to the Heterogeneity-Aware Transformer (Section~\ref{method:DRoPE}), the Mixture-of-Size Encoder adaptively tokenizes the raw time series $\mathbf{X}$ through a three-stage process.
First, the series is partitioned into fixed-size \textit{segments}, where a router dynamically selects optimal temporal granularities \textit{for each segment}.
Next, for each activated granularity, the segment is simultaneously patchified and projected into the latent space by the expert network corresponding to that granularity.
Finally, these multi-granularity embeddings are fused into a unified token sequence representing the input series.

Notably, the Mixture-of-Size Encoder is both \textit{sparse} and \textit{dynamic}. 
A gating mechanism equipped with null experts (described below) adaptively selects relevant granularities \textit{for each segment}, enabling sparse expert activation and segment-specific granularity adaptation.

\minisection{Coarse Segmentation}
As shown in Figure~\ref{fig:intro}(b) in Section~\ref{sec:intro} where time series exhibits inherent temporal heterogeneity, we first apply coarse segmentation to the input time series $\mathbf{X}$ and process each segment independently in subsequent tokenization and encoding stages.
Specifically, $\mathbf{X}$ is partitioned into a sequence of non-overlapping coarse segments $\mathbf{X} = [\mathbf{x}_1, \ldots, \mathbf{x}_N]$,
where each segment $\mathbf{x}$ has a fixed length $P$ and the total number of segments is $N=\lceil {T}/{P} \rceil$.
In the following, we operate within each segment $\mathbf{x}$ and therefore omit the segment index for notational simplicity.

\minisection{Granularity}
For a time series segment $\mathbf{x}$, we adopt a multi-granularity perspective. 
\textit{Granularity} refers to the temporal resolution at which the segment is tokenized and encoded, as defined below.
\begin{definition}[Granularity Level]
\itshape
A granularity level $i \in \{1, \ldots, G\}$ is defined by 
a specific patch size $P_i$ and its corresponding encoding network $\pi_i$, 
under which the time series segment $\mathbf{x}$ is patchified and mapped into a latent representation.
\end{definition}
Intuitively, finer granularity levels capture high-frequency local variations, while coarser granularity levels emphasize low-frequency global trends. 
At each granularity level $i$, the patchification and embedding operations described below are applied independently and in parallel, producing representations that characterize the segment at the corresponding temporal resolution.

\minisection{Step 1: Top-$K$ Granularity Routing}
Applying encoding at an appropriate granularity level for each time series segment is non-trivial due to the inherent temporal heterogeneity of time series data. 
To address this challenge, we introduce a dynamic routing mechanism equipped with a lightweight \textit{router} that adaptively selects an optimal subset of granularity levels for subsequent encoding of a given segment.
Inspired by the Mixture-of-Experts (MoE) paradigm~\citep{jacobs1991adaptive, shazeer2017outrageously}, we instantiate $G$ granularity-specific experts, each corresponding to a distinct granularity level, together with $Z \in \{1, \dots,K-1\}$ computation-free null experts. 
The inclusion of null experts allows the model to implicitly skip encoding at certain granularities, enabling adaptive control over the \textit{effective number of active granularities for each input segment}.
For each expert $i \in \{1, \dots, G+Z\}$, the routing weights are computed via a learnable affinity score $s_i = \mathbf{w}_i^\top \mathbf{x}$, normalized with a bias-corrected softmax $\tilde{s}_i = \left( \exp(s_i + \boldsymbol{b}_i) \right) / \left( \sum_{j=1}^{G+Z} \exp(s_j + \boldsymbol{b}_j) \right),$ where $\mathbf{w}_i$ denotes learnable parameters and $\boldsymbol{b}_i$ is a bias for auxiliary-loss-free load balancing~\citep{liu2024deepseek} (details in Appendix~\ref{app:bias}).
The router then activates the indices $\mathcal{I}_{\mathrm{valid}} = \operatorname*{arg top\text{-}K}_{i \in [G+Z]} ~ \tilde{s}_i \cap \{1, \dots, G\}$.
By allowing null experts to dynamically prune the number of active experts, \method dynamically adapts the effective resolution to local information while maintaining a sparse, parameter-efficient architecture.

\minisection{Step 2: Granularity-wise Encoding}
For each selectively activated granularity expert at level $i$, we apply the corresponding patching and embedding operations.
Specifically, a given time series \emph{segment} $\mathbf{x}$ is re-partitioned into a sequence of patches at granularity $i$:
\begin{equation}
    [ \boldsymbol{p}_{i,1}, \dots, \boldsymbol{p}_{i,M_i} ] = \mathrm{Patchify}(\mathbf{x}, i) \in \mathbb{R}^{M_i \times P_i},
\end{equation}
where $P_i$ denotes the patch size at granularity $i$, and
$M_i = \left\lceil {P}/{P_i} \right\rceil$ is the resulting number of patches obtained via non-overlapping segmentation, with zero-padding applied when necessary.
Each patch is then projected into the latent space to obtain an embedding sequence
$\boldsymbol{E}_{i} = [ \boldsymbol{e}_{i,1}, \dots, \boldsymbol{e}_{i,M_i} ]$:
\begin{equation}
    \quad \boldsymbol{e}_{i,m} = \pi_i(\boldsymbol{p}_{i,m}) \in \mathbb{R}^{D_h}, m\in[1,M_i] ~.
    \label{eq:granularity_embedding}
\end{equation}
where $\pi_i$ denotes a granularity-specific projection function, implemented as a two-layer MLP.

\textbf{Step 3: Mixture-of-Size Fusion.}
In the final encoder stage, we fuse the embeddings produced by the activated granularity experts into a unified token sequence, serving as the input to the Heterogeneity-Aware Transformer.
First, for a time-series segment $\mathbf{x}$, we align all granularity-specific embeddings $\boldsymbol{E}_i$ to the finest activated resolution
$
P_{\min} = \min \{ P_i \mid i \in \mathcal{I}_{\mathrm{valid}} \},
$
yielding a target sequence length $M = \lceil P / P_{\min} \rceil$.
This alignment is crucial to eliminate representation bias arising from unequal sequence lengths induced by different patch sizes across granularities.
To enable efficient alignment, we design the patch sizes to be nested, i.e., $P_{i+1}$ is divisible by $P_i$.
Under this constraint, each embedding sequence $\boldsymbol{E}_i \in \mathbb{R}^{M_i \times D_h}$ can be deterministically expanded to the target resolution via repetition (details and an example in Appendix~\ref{app:expansion_def}):
$\tilde{\boldsymbol{E}}_i = \mathrm{Expand}(\boldsymbol{E}_i) \in \mathbb{R}^{M \times D_h}.
$
The fused representation is obtained by a gated weighted summation over the activated granularities:
$
    \bar{\boldsymbol{E}}
    = \sum_{i \in \mathcal{I}_{\mathrm{valid}}}
    c_i \, \tilde{\boldsymbol{E}}_i,
    c_i = \tilde{s}_i / {\sum_{j \in \mathcal{I}_{\mathrm{valid}}} \tilde{s}_j}.
$

\textbf{Sample-level Result.}
Finally, the representations from all coarse segments  
are concatenated to form the encoder output
\begin{equation}
    \boldsymbol{H}
    = \mathrm{Concat}(\bar{\boldsymbol{E}}_1, \ldots, \bar{\boldsymbol{E}}_N)
    \in \mathbb{R}^{\tilde{T} \times D_h},
\end{equation}
where $\tilde{T} = \sum_{n=1}^{N} M_n$ and $M_n$ is the embedding sequence length of the $n$-th segment, forming the representation sequence for the given time series sample $\mathbf{X}$.

Unlike prior approaches such as \citet{chen2024pathformer, zhang2024elastst}, which impose a \emph{static} multi-scale partitioning over the entire time series, our method (i) dynamically adapts the tokenization granularity at the \emph{segment level}.
(ii) decouples temporal heterogeneity from a fixed representational scale by explicitly and dynamically assigning patch size via routing and multi-granularity encoding, enabling more effective \textit{intra-patch representation learning}.
This adaptive behavior cannot be achieved via globally predefined scales applied statically and is empirically validated in Section~\ref{sec:mosdp_analysis} and Appendix~\ref{app:segment_level_adaptation}.
As a result, varying temporal patterns are modeled via the lightweight Mixture-of-Size Encoder, reducing the burden of subsequent model parameterization.

\subsection{Heterogeneity-Aware Transformer}
\label{method:DRoPE}

Following Mixture-of-Size Encoder, the fused representation $\boldsymbol{H} \in \mathbb{R}^{\tilde{T} \times D_h}$ is processed by a Heterogeneity-Aware Transformer, a stack of standard Transformer \citep{vaswani2017attention} encoder blocks, yielding representations $\boldsymbol{H}^{\text{en}} \in \mathbb{R}^{\tilde{T} \times D_h}$ that are subsequently fed to the Multi-Patch Decoder.
To inject temporal information into the self-attention mechanism, we propose Dynamic Rotary Position Embedding (DRoPE). 
Unlike static encodings such as~\citet{vaswani2017attention,su2024roformer}, DRoPE adaptively calibrates temporal structure to account for instance-specific periodicities and the varying physical durations of dynamic patches.

\minisection{Base Structure of DRoPE}
Our dynamic position encoding is based on the standard Rotary Position Embedding (RoPE)~\cite{su2024roformer}, where a query or key vector $\boldsymbol{z} \in \mathbb{R}^{D_{\mathrm{rot}}}$, with rotation dimension $D_{\mathrm{rot}}$, at token index $t \in [1, \tilde{T}]$ is treated as a set of $D_{\mathrm{rot}}/2$ complex numbers.
Each pair is rotated by an angle proportional to the position $t$ and a fixed frequency $\theta_d$ as below, where $d \in [0, D_{\mathrm{rot}}/2-1]$ indexes the dimension pair:
\begin{equation}
    f_{\text{RoPE}}(\boldsymbol{z}, t) = (\boldsymbol{z}_{2d} + i \boldsymbol{z}_{2d+1}) e^{i t \theta_{d}}
\end{equation}
Here the frequencies are traditionally pre-defined as $\theta_d = b^{-2d/D_{\mathrm{rot}}}$.
DRoPE generalizes this formulation by dynamically transforming both the frequency $\theta$ and the token position $t$.
A detailed derivation of the RoPE calculation is provided in Appendix~\ref{app:DRoPE}. 

\minisection{Instance-Level Frequency Modulation}
Standard RoPE uses a uniform frequency $\theta$ for all sequences, failing to capture the unique periodic structures of heterogeneous time series (analysis in Theorem~\ref{thm:rope-periodic}). 
We propose to modulate the base frequencies using instance-specific spectral features.
Specifically, given a time series input $\mathbf{X}$, we extract the magnitudes of the first $\omega$ low-frequency components via Fast Fourier Transform, denoted as $\mathbf{X}_\mathrm{fft}$.
A lightweight MLP maps these features to a layer-specific scaling factor $\gamma$ and bias $\beta$ to produce \emph{adaptive frequencies} $\theta_\mathrm{inst}$.
For brevity, we omit the layer index $l$ in this section.
To accommodate the exponentially decaying nature of RoPE's base frequencies, we perform this affine transformation in log-space:
\begin{equation} 
\log \theta_{\text{inst},d} = \gamma_d \cdot \log \theta_{d} + \beta_d ~, \text{where~} \gamma,\beta=\mathrm{MLP}(\mathbf{X}_\mathrm{fft})
\end{equation}
This log-space adaptation allows \method to robustly re-scale the temporal rotation based on the dominant information like trends and seasonalities in the specific time series instance (see Appendix~\ref{app:drope_details} for a detailed discussion).

\minisection{Granularity-Aware Position Calibration}
In our multi-granularity framework, the discrete token index $t$ is no longer an accurate proxy for time because different tokens represent different physical durations, a direct consequence of the granularity routing and tokenization in our Mixture-of-Size Encoder, as described in Section~\ref{sec:mos-encoder}.
To maintain temporal consistency across mixed patch sizes, we replace the index $t$ with a calibrated position $\tilde{t}=\sum_{k=1}^{t-1} {\bar{P}_k}/{\bar{P}_\mathrm{g\_min}}$, where $\bar{P}_k$ is the patch size of the $k$-th token and $\bar{P}_\mathrm{g\_min}=\min\{P_i\}_{i=1}^G$ is the globally minimum patch size of the finest granularity.

Unifying these two components, rotating the hidden vector $\boldsymbol{z}$ at position $t$ as follows:
\begin{equation}
    f_{\text{DRoPE}}(\boldsymbol{z}, t) = (\boldsymbol{z}_{2d} + i \boldsymbol{z}_{2d+1}) e^{i \cdot \tilde{t} \cdot \theta_{\text{inst},d}}
\end{equation}
By measuring relative distances in terms of absolute physical time and instance-specific frequencies, DRoPE enables the Transformer to learn consistent temporal dependencies considering the diverse encoding granularity and the intrinsic properties of the data sample.

\subsection{Multi-Patch Decoder}
\label{sec:multi_patch_prediction}
To mitigate the error accumulation typical of patch-wise autoregressive decoding while maintaining structural flexibility, \method utilizes a Multi-Patch Decoder. 
This component forecasts $J$ future patches in parallel, effectively decoupling the prediction horizon from iterative dependencies within each autoregressive round.
We first introduce a sequence of $J$ learnable forecast tokens, where each $\mathbf{q}_j \in \mathbb{R}^{D_h}$ has the hidden dimension $D_h$, and stack them into $\mathbf{Q}_{\text{forecast}} = [\mathbf{q}_1, \dots, \mathbf{q}_J]^\top \in \mathbb{R}^{J \times D_h}$.
Multi-Patch Decoder processes these token embeddings with the latent representations $\boldsymbol{H}^{\text{en}} \in \mathbb{R}^{\tilde{T} \times D_h}$ from the above Heterogeneity-Aware Transformer through cross-attention, to produce a sequence of latent forecasting embeddings $\boldsymbol{h}^{\text{de}} \in \mathbb{R}^{J \times D_h}$. 
    
Each embedding $\boldsymbol{h}^{\text{de}}_j$ is then projected into the observation space via a shared prediction head, which is implemented as a residual FFN block~\cite{das2024decoder}, to generate a future patch of fixed length $H_s$. Within one decoding round, the prediction is
\begin{equation}
    \hat{\mathbf{Y}}_{(j-1)H_s+1: jH_s} = \text{ResidualFFN}(\boldsymbol{h}^{\text{de}}_j),\quad j \in [1,J]
\end{equation}

This multi-token design serves two primary objectives: (i) \emph{error mitigation and computational efficiency} via parallel patch prediction, and (ii) \emph{horizon flexibility} for supporting variable prediction horizons.
We provide a detailed comparative analysis of this design in Appendix~\ref{app:multi_patch_analysis}.
By pairing this flexible Multi-Patch Decoder with the adaptive Mixture-of-Size Encoder, \method achieves a unified and parameter-efficient framework for zero-shot time series forecasting.

\subsection{Training Corpus}
For pre-training, we curated the Predictability-Stratified Time Series (PreSTS) corpus, comprising over 300 billion time points.
PreSTS integrates large-scale real-world time series from diverse domains and a complementary synthetic dataset to ensure broad coverage.
All training, validation, and test splits from the GIFT-Eval~\citep{aksu2024gifteval} and TSLib~\citep{wang2024tssurvey} benchmarks are explicitly excluded to prevent data leakage.
While dataset diversity is widely regarded as essential for TSFM training, recent studies indicate that low-predictability or anomalous sequences can substantially impair forecasting~\citep{chengrobusttsf}. 
To address this, we stratify real-world datasets into five tiers based on predictability and assign higher sampling probabilities to more predictable data during training.
This strategy promotes stable and high-quality pre-training without compromising data diversity.
Further details on dataset construction, along with an experimental analysis decoupling the contribution of the corpus from our model architecture, are provided in Appendix~\ref{app:data} and Appendix~\ref{app:ablation_data}, respectively.

\section{Experiment}
\label{sec:exp}
In this section, we assess the performance of \method and its components by addressing three research questions.
\textbf{RQ1:} 
Does \method achieve superior zero-shot generalization capabilities while maintaining parameter efficiency compared to previous methods?
(Section \ref{sec:zero_shot_eval})
\textbf{RQ2:} Does the Mixture-of-Size Encoder effectively handle time series heterogeneity by adapting its granularity based on local information density? (Section \ref{sec:ablation_study} and \ref{sec:mosdp_analysis})
\textbf{RQ3:} Does DRoPE effectively generate customized position embeddings tailored to the unique temporal structure of each time series instance? (Section \ref{sec:ablation_study} and~\ref{exp:DRoPE_analysis})

\subsection{Experiment Settings}
\label{sec:eval_datasets_and_metric}
To comprehensively evaluate our proposed method, we conducted zero-shot assessments on two well-known public benchmarks: GIFT-Eval~\citep{aksu2024gifteval} and Time-Series-Library (TSLib)~\citep{wang2024tssurvey}.
We train \method in three sizes: mini (10M), small (23M) and base (53M).
The training details are provided in Appendix \ref{app:training_details}. 

\begin{table*}[htbp]
  \centering
  \caption{
  Performance evaluation on the GIFT-Eval.
  We evaluate using normalized MASE and CRPS (Section \ref{sec:eval_datasets_and_metric}).
  Baselines encompass statistical, Deep Learning (DL), and TSFM methods, with TSFMs categorized into TestData Leakage and Zero-Shot following official protocols.
  TestData Leakage signifies that the model's training set included test data, whereas Zero-Shot indicates that the model was trained without any data from the test set or its corresponding training split.
  Baseline results are officially reported by GIFT-Eval.
  }
  \label{tab:zero-shot_gift}
  \resizebox{\textwidth}{!}{
  \begin{tabular}{lcccccccccccccc}
    \toprule
    \multicolumn{1}{l}{\textbf{Type}} & \multicolumn{1}{c}{\textbf{Statistical}} & \multicolumn{2}{c}{\textbf{DL (Full-Shot)}} & \multicolumn{4}{c}{\textbf{TSFMs (TestData Leakage)}}
    & \multicolumn{7}{c}{\textbf{TSFMs (Zero-Shot)}} \\
    \cmidrule(lr){1-1}
    \cmidrule(lr){2-2} \cmidrule(lr){3-4} \cmidrule(lr){5-8} \cmidrule(lr){9-15}
    \multicolumn{1}{l}{\textbf{Method}} & \makecell{Seasonal\\Naïve} & \makecell{DLinear} & PTST. & TTM & $\text{Chronos}$ & \makecell{Chronos\\Bolt} & $\text{TimesFM}$ & $\text{Moirai}$ & VisionTS & $\text{Ying.}$ & $\text{Toto}$ & $\text{Sundial}$ & \makecell{$\method_{s}$\\\textbf{(ours)}} & \makecell{$\method_{b}$\\\textbf{(ours)}} \\
    \multicolumn{1}{l}{\textbf{\#Params}} & - & - & - & 5M & 709M & 205M & 500M & 311M & 112M & 300M & 151M & 128M & 23M & 53M \\
    \midrule
    \multicolumn{1}{l}{\textbf{MASE} $\downarrow$} & 1.000 & 1.061 & 0.849 & 1.020 & 0.870 & 0.808 & 0.758 & 0.875 & 0.863 & 0.798 & 0.750 & 0.750 & \underline{0.748} & \textbf{0.738} \\
    \multicolumn{1}{l}{\textbf{CRPS} $\downarrow$} & 1.000 & 0.846 & 0.587 & 0.873 & 0.574 & 0.574 & 0.550 & 0.599 & 0.755 & \underline{0.548} & \textbf{0.517} & 0.559 & 0.554 & \underline{0.548} \\
    \bottomrule
  \end{tabular}%
  }
\end{table*}

\begin{figure*}[h]
\centering
\includegraphics[width=\textwidth]{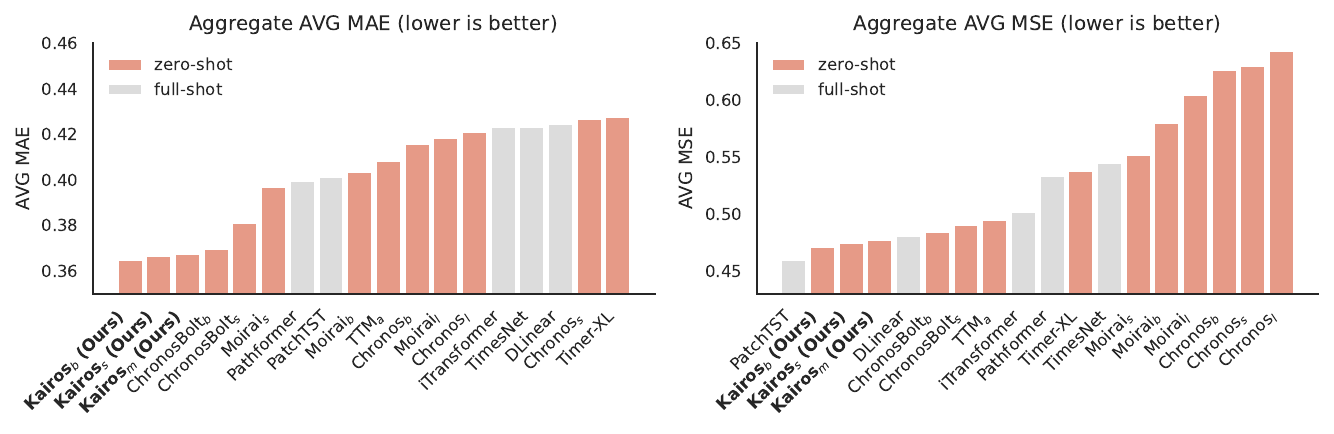}
\caption{
Zero-shot performance on TSLib averaged over prediction lengths \{96, 192, 336, 720\}.
The subscripts $l$, $b$, $s$, $m$, and $a$ represent model sizes of large, base, small, mini and advanced, respectively.
The complete experimental results are presented in Appendix \ref{sec:additional_results}.}
\label{fig:zero-shot_tslib}
\end{figure*}

\minisection{Benchmarks}
The GIFT-Eval benchmark comprises 97 tasks spanning short- (55), medium- (21), and long-term (21) horizons, facilitating a comprehensive assessment.
For TSLib, we selected the ETT and Weather~\citep{zhou2021informer} datasets, consistent with Time-MoE~\citep{shi2024timemoe}.
To test adaptability to diverse frequencies, we included the Saugeen datasets~\citep{godahewa2021monash}, denoted as Saugeen (D) and Saugeen (W) for daily and weekly intervals, respectively.
Details of the compared methods, evaluation datasets, splitting configurations, and evaluation length selection are provided in Appendices \ref{app:compared_methods}, \ref{app:eval_data}, and \ref{app:Evaluation_Length_Selection}.

\minisection{Metric}
In our evaluation of GIFT-Eval, consistent with prior research~\citep{auer2025tirex, liu2025sundial}, we employ the official Mean Absolute Scaled Error (MASE) and Continuous Ranked Probability Score (CRPS) metrics to assess point and probabilistic forecasting performance, respectively.
These metrics are first normalized by the Seasonal Naïve baseline for each individual task.
The final score is then computed as the geometric mean of these normalized values.
For the TSLib benchmark, we followed previous works~\citep{liu2023itransformer, liu2024timer, nie2023time} by adopting Mean Squared Error (MSE) and Mean Absolute Error (MAE) to evaluate time series forecasting performance. 
The metric calculations are detailed in Appendix \ref{app:eval_metric}.

\subsection{Zero-shot Evaluation}
\label{sec:zero_shot_eval}
\label{sec:gift-eval}
\textbf{Finding 1:}
\textit{\method achieves superior zero-shot forecasting performance with significantly higher parameter efficiency and learns highly transferable representations..}
On the GIFT-Eval benchmark, as presented in Table~\ref{tab:zero-shot_gift}, $\method_{\text{base}}$ (53M) achieves the best MASE and the second-best CRPS among state-of-the-art task-specific deep learning methods and other TSFMs.
Notably, $\method_{\text{small}}$ (23M) surpasses both Toto and Sundial in MASE, despite having a parameter count that is 6.6 and 5.6 times smaller, respectively. 
On the TSLib benchmark, as shown in Figure \ref{fig:zero-shot_tslib}, our lightweight $\method_{\text{mini}}$ (10M) surpasses the performance of both recent advanced TSFMs and the majority of full-shot deep learning models.
These results confirm that \method delivers a significant improvement in predictive capability without relying on massive scale (statistical analysis is provided in Appendix~\ref{app:statistical_analysis}).
Beyond zero-shot forecasting, \method-Base outperforms full-shot PatchTST and DLinear after fine-tuning on just 1\% of training windows (Appendix~\ref{app:few_shot}).
Furthermore, we demonstrate in Appendix~\ref{app:classification_results} that these learned representations are highly transferable to downstream classification tasks.

\subsection{Ablation Study}
\label{sec:ablation_study}
\textbf{Finding 2:}
\textit{The synergistic integration of Mixture-of-Size Encoding, DRoPE, and the Multi-Patch Decoding is critical for adapting to heterogeneous time series and achieving superior forecasting performance.}
Table~\ref{tab:ablation} details the component-wise contributions.
(i) \textit{Encoder Design}: \method outperforms variants using fixed patch sizes or excluding null experts, confirming that dynamic, sparse multi-granularity modeling is critical for capturing diverse temporal patterns.
(ii) \textit{Positional Embedding}: DRoPE proves superior to standard RoPE~\citep{su2024roformer}. Our ablation further isolates the gains from \textit{Instance-Level Frequency Modulation} and \textit{Granularity-Aware Position Calibration}, verifying the benefit of adapting to both spectral heterogeneity and physical time distortion caused by dynamic patching.
(iii) \textit{Decoding Strategy}: The Multi-Patch Decoding shows clear superiority over standard single-patch autoregressive decoding. 
The integration yielded the best performance across 97 GIFT-Eval tasks, which validates the superiority of our designs.
Furthermore, the performance gap widens as the forecast horizon increases, demonstrating \method's superior long-term forecasting capability.

\begin{table}[htbp]
\caption{Ablation study comparing encoder design, positional embedding and decoder strategy. 
Models were evaluated using the normalized MASE (detailed in Section \ref{sec:eval_datasets_and_metric}) across prediction horizons and as an aggregate across all tasks.}
\label{tab:ablation}
\centering
\resizebox{0.8\columnwidth}{!}{%
\begin{tabular}{lcccc}
\toprule
\textbf{Model Variants} & \textbf{Short} & \textbf{Medium} & \textbf{Long} & \textbf{AVG} \\
\midrule
\multicolumn{5}{l}{\textit{\textbf{Impact of Encoder Design}}} \\
\quad w/ Fixed Patch Size (32)  & 0.724 & 0.802 & 0.820 & 0.761\\
\quad w/ Mixture-of-Size (w/o Null Experts) & 0.720 & 0.770 & 0.800 & 0.748 \\
\midrule
\multicolumn{5}{l}{\textit{\textbf{Impact of Positional Embedding}}} \\
\quad w/ Standard RoPE (Fixed $\theta$) & 0.729 & 0.807 & 0.835 & 0.767 \\
\quad w/ Only Instance-Level Frequency Modulation & 0.719 & 0.767 & 0.797 & 0.746 \\
\quad w/ Only Granularity-Aware Position Calibration & 0.727 & 0.796 & 0.807 & 0.758 \\
\midrule
\multicolumn{5}{l}{\textit{\textbf{Impact of Decoder Strategy}}} \\
\quad w/ Single-Patch Autoregressive & \textbf{0.705} & 0.794 & 0.848 & 0.753 \\
\midrule
\rowcolor{gray!10} \textbf{\method (Full Model)} &  0.709 & \textbf{0.761} & \textbf{0.794} & \textbf{0.738} \\
\bottomrule
\end{tabular}%
}
\end{table}

\subsection{Model Analysis}
\subsubsection{Mixture-of-Size Tokenization Analysis}
\label{sec:mosdp_analysis}

\begin{wrapfigure}{r}{0.55\textwidth} 
\centering
\vspace{-12pt}
\includegraphics[width=0.55\columnwidth]{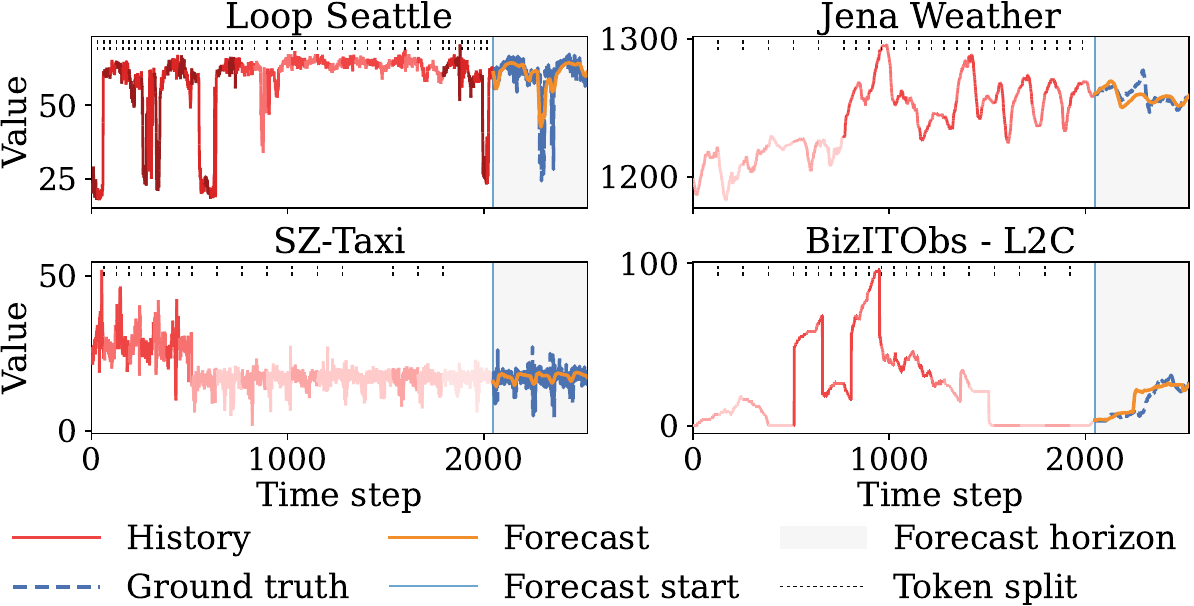}
\caption{Adaptive tokenization visualization. Tokens are demarcated by dotted lines and varying colors, where \textcolor{red!80!black}{\textbf{deeper red}} curves indicate finer granularity. \method adapts granularity to information density, applying finer processing to volatile regions.
}
\vspace{-8pt}
\label{fig:Dataset-Specific-Preferences}
\end{wrapfigure}

\textbf{Finding 3:} 
\textit{\method demonstrates adaptive capability to varying information densities, enabling it to model complex temporal dynamics with appropriate granularity.}
Instead of enforcing a uniform resolution, \method dynamically routes segments to optimal granularity based on local information.
As visualized in Figure~\ref{fig:Dataset-Specific-Preferences}, \method strategically allocates fine-grained tokens to regions with abrupt changes or high volatility, while employing coarse-grained tokens for stable trends. 
This adaptive allocation mechanism reduces computational redundancy on low-information segments, allowing the model to achieve superior performance with a compact parameter budget compared to rigid architectures.
See Appendix~\ref{app:infer_speed} for a detailed inference speed analysis.
Furthermore, to demonstrate that \method learns genuine segment-level adaptation rather than merely relying on a generic local multi-scale inductive bias, we provide additional experimental evidence in Appendix~\ref{app:segment_level_adaptation}.

\subsubsection{DRoPE Analysis}
\label{exp:DRoPE_analysis}
\begin{wrapfigure}{r}{0.55\textwidth} 
\centering
\vspace{-12pt}
\captionof{table}{
Causal analysis of adaptive modulation by DRoPE. We evaluate the impact of disrupting or removing the instance-wise $\theta$ adjustment. 
Degradation indicates relative performance drop compared to DRoPE.
}
\label{tab:drope_analysis}
\resizebox{0.55\columnwidth}{!}{
\begin{tabular}{lcccc}
\toprule
\textbf{Method} & \textbf{DRoPE} & \textbf{Intra-Dataset} & \textbf{Standard RoPE} & \textbf{Inter-Dataset} \\
 & \textbf{(Ours)} & \textbf{Shuffle} & \textbf{(Fixed $\theta$)} & \textbf{Shuffle} \\
\midrule
MASE $\downarrow$ & \textbf{0.738} & 0.751 & 0.767 & 0.947 \\
Degradation & -- & 1.76\% & 3.93\% & 28.32\% \\
\bottomrule
\vspace{-25pt}
\end{tabular}
}
\end{wrapfigure}
\textbf{Finding 4:} \textit{Modulating RoPE frequencies $\theta$ instance-wise can indeed better model time series temporal relationships.}
To verify that DRoPE's performance gains stem from its instance-specific $\theta_{\text{inst}}$ modulation, we designed control experiments that disrupt this mechanism (Table~\ref{tab:drope_analysis}; Appendix~\ref{sec:DRoPE_analysis}): 
(i) Intra-Dataset Shuffle: randomly permute $\theta_{\text{inst}}$ across instances within the same dataset;
(ii) Inter-Dataset Shuffle: assign $\theta_{\text{inst}}$ from instances of different datasets; 
(iii) Fixed RoPE: use standard RoPE without any modulation.
The results reveal a clear performance gradient (DRoPE $\succ$ Intra-Shuffle $\succ$ Fixed RoPE $\succ$ Inter-Shuffle), indicating that DRoPE learns instance-specific representations to modulate $\theta$. 
Intra-Dataset Shuffle outperforms Fixed RoPE, as samples within the same dataset share similar spectral characteristics, allowing shuffled $\theta_{\text{inst}}$ to remain partially compatible.
However, Intra-Dataset Shuffle still incurs a 1.76\% degradation compared to DRoPE, suggesting that instance-level adaptation contributes to the performance gains beyond dataset-level statistics.
Inter-Dataset Shuffle yields the worst results (28.32\% degradation), as $\theta_{\text{inst}}$ from different datasets cannot capture the temporal dynamics of the target series.
These results collectively validate the effectiveness of instance-level $\theta_{\text{inst}}$ adaptation in DRoPE.

\section{Conclusions}

We introduce \method, a parameter-efficient foundation model explicitly tailored for time series heterogeneity. Its architecture features three key innovations: a Mixture-of-Size Encoder for adaptive granularity, a Heterogeneity-Aware Transformer with Dynamic RoPE for instance-specific temporal calibration, and a Multi-Patch Decoder for efficient parallel forecasting. Extensive experiments demonstrate that \method achieves superior performance, validating the necessity of aligning architectural inductive biases with the structural nuances of time series data.
One limitation of our current approach is that \method relies on channel-independent modeling~\citep{nie2023time} to support multivariate time series and does not explicitly capture inter-variable dependencies. 
Future work aims to model these interactions and extend \method to tasks beyond forecasting.

\begin{ack}
The research was supported by the National Natural Science Foundation of China (Grant No. 62406193) and the ShanghaiTech AI Initiative (Grant No. AI2026B08).
The authors also gratefully acknowledge assistance from the Key Laboratory of Intelligent Perception and Human-Machine Collaboration (ShanghaiTech University), Ministry of Education, and the HPC Platform of ShanghaiTech University.
There are no competing interests to declare.
\end{ack}

\bibliographystyle{plainnat}
\bibliography{main}

\newpage
\appendix

\section{Limitations and Future Work}
\label{app:limitation}

While \method demonstrates superior parameter efficiency and zero-shot performance, we acknowledge several limitations that provide avenues for future research.

\textbf{Channel-Independent Modeling.} 
Currently, \method adopts a channel-independent modeling approach~\citep{nie2023time} to handle multivariate time series, treating each variable as an individual sequence without explicitly modeling the relationships between variables. 
However, our architectural innovations are orthogonal to the channel-independence assumption. 
It is easy to incorporate channel dependency modeling for further enhancement.
For instance, a channel-mixer block can be easily integrated to explicitly capture cross-channel correlations. 
We plan to explore these extensions in future work.

\textbf{Evaluation Scope.} 
Our empirical evaluation primarily focuses on zero-shot forecasting, which is the primary evaluation setting for TSFMs~\citep{liu2025sundial, cohen2025time}.
As a generative task, forecasting directly reflects whether a model can capture temporal structures such as trend, seasonality, and local dynamics.
Furthermore, we provide evaluation results on classification tasks in Appendix~\ref{app:classification_results}, which demonstrate that \method successfully learns transferable representations.
We plan to further extend \method to tasks such as anomaly detection and imputation in future versions.

\textbf{Probabilistic Forecasting.}
Although \method-Base achieves the best normalized MASE in Table~\ref{tab:zero-shot_gift}, its normalized CRPS of 0.548 is higher than Toto's 0.517.
\method-Base uses 53M parameters compared with Toto's 151M, offering a favorable accuracy--model-size trade-off, but a gap remains in probabilistic forecasting performance.
Improving probabilistic accuracy while retaining this parameter efficiency is a direction for future work.

\section{Etymology of \method}
\label{app:name}
\method was chosen for its mythological significance as the Greek god of the ``right or critical moment'', reflecting our model's ability to select the ideal set of granularities and positional encoding for time series characterized by varying information densities and frequencies. 
Furthermore, this continues a thematic tradition seen in other foundational models we benchmark against, such as Chronos~\citep{ansari2024chronos} (the personification of time) and Moirai~\citep{woo2024moirai} (the personifications of destiny).

\section{Transferability on Classification Tasks}
\label{app:classification_results}

To verify whether \method learns generalizable features that transfer beyond forecasting tasks, we evaluate its performance on downstream time series classification.

\textbf{Experimental Setup.} Following MOMENT~\citep{goswami2024moment}, we evaluate on 91 datasets from the UCR archive~\citep{dau2019ucr}. To isolate representation quality, \method-Base and MOMENT use a frozen encoder with a lightweight classifier, comparing them against fully fine-tuned baselines (GPT4TS~\citep{zhou2023one}, TimesNet~\citep{wu2022timesnet}).

\textbf{Results.} The classification results are summarized in Table~\ref{tab:ucr_classification}. 
Despite relying solely on frozen representations, \method consistently outperforms both MOMENT and the fully fine-tuned baselines. These results provide evidence that our architectural innovations do not merely overfit to forecasting dynamics.
Instead, by addressing temporal heterogeneity, \method extracts high-level semantic features that are highly transferable to adaptation-heavy settings like classification. 
This solidifies \method's capacity as a general-purpose foundation model capable of supporting diverse downstream time series applications.

\begin{table}[htbp]
\centering
\caption{Representation transfer results on 91 UCR classification datasets. Baseline results follow MOMENT~\citep{goswami2024moment}.}
\label{tab:ucr_classification}
\begin{tabular}{lccc}
\toprule
\textbf{Model} & \textbf{Accuracy $\uparrow$} & \textbf{Rank Mean $\downarrow$} & \textbf{Rank Std $\downarrow$} \\
\midrule
MOMENT (Frozen) & 0.794 & 1.90 & 0.80 \\
GPT4TS (Fine-tuned) & 0.567 & 3.27 & 0.81 \\
TimesNet (Fine-tuned) & 0.573 & 3.32 & 0.74 \\
\midrule
\rowcolor{gray!10} \textbf{\method-Base (Ours, Frozen)} & \textbf{0.818} & \textbf{1.51} & \textbf{0.71} \\
\bottomrule
\end{tabular}
\end{table}

\section{Analysis of Multi-Patch Prediction}
\label{app:multi_patch_analysis}

Unlike task-specific, non-autoregressive models such as TFT~\citep{lim2021temporal} and Informer~\citep{zhou2021informer}, which predict all future time points simultaneously, their architecture confines them to forecasting a fixed length identical to the predefined prediction horizon used during training, precluding variable-length predictions during inference. 
Conversely, TSFMs typically employ an iterative process: they predict a segment of the sequence, append it to the historical data, and then forecast the subsequent segment, thereby enabling autoregressive predictions of arbitrary length.
To mitigate the cumulative errors inherent in this autoregressive methodology, \method introduces forecast tokens designed to predict multiple patches within each autoregressive step.

As briefly mentioned in Section~\ref{sec:multi_patch_prediction}, the multi-token design of our Multi-Patch Decoder is strategically motivated by two limitations in predominant forecasting paradigms:
(i) \emph{Error mitigation and efficiency}: Compared to conventional single-step autoregressive models~\citep{ansari2024chronos,liu2024timer}, predicting $J$ patches 
in parallel
improves computational efficiency and mitigates cumulative error by reducing iterative decoding steps.
(ii) \emph{Horizon flexibility}:
Unlike models that generate a single larger patch~\citep{das2024decoder}, our approach 
offers superior granularity for variable prediction horizons.
By assembling the forecast from the specific subset of tokens required, \method ensures high-fidelity alignment with the target length and avoids the imprecise truncation associated with oversized outputs.

\begin{figure}[htbp]
\centering
\includegraphics[width=0.7\columnwidth]{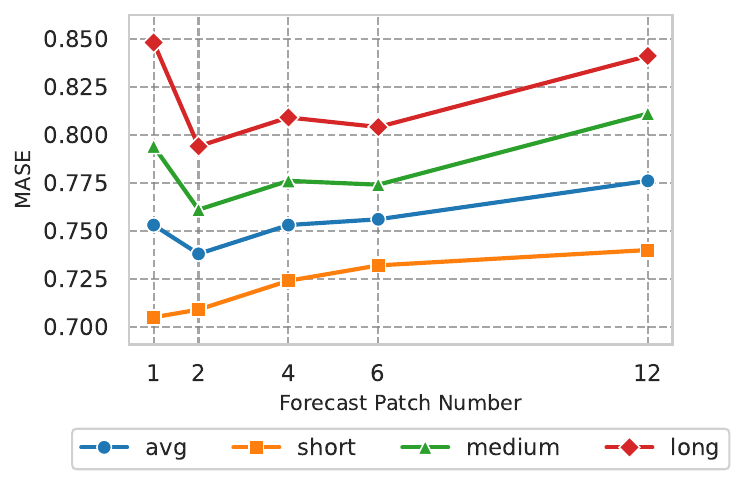}
\caption{Performance analysis of multi-patch prediction on the GIFT-Eval benchmark across short, medium, and long horizons. 
We vary the number of forecast patches and find that a forecast patch number of 2 achieves the optimal trade-off, resulting in the best overall performance (lowest normalized MASE).}
\label{fig:patch_number_performance}
\end{figure}
\minisection{Effect of Forecast-Token Count}
To empirically validate this design and determine the optimal setting, we present a detailed analysis of the multi-patch prediction strategy. During the training phase, we experimented with a range of forecast patch numbers, specifically $J=\{1, 2, 4, 6, 12\}$.
The corresponding evaluation results on the GIFT-Eval benchmark are illustrated in Figure~\ref{fig:patch_number_performance}.
Our observations reveal that when $J=1$, which corresponds to the conventional approach of predicting a single patch~\citep{ansari2024chronos,liu2024timer,das2024decoder}, the model achieves optimal performance in short-term forecasting.
However, this method necessitates multiple iterations of autoregressive prediction, leading to a significant degradation in performance for medium- and long-term forecasting.

Conversely, as we increase the forecast patch number $J$, the number of required autoregressive steps is markedly reduced.
For instance, when $J=2$, the autoregression frequency for medium- and long-term predictions is halved compared to the $J=1$ case.
This reduction yields a substantial improvement in forecasting accuracy over these longer horizons, demonstrating that our proposed multi-patch prediction strategy effectively mitigates the cumulative error inherent in autoregressive processes for medium- and long-term forecasting.

Nevertheless, we noted that the forecasting performance does not improve indefinitely with an increasing the forecast patch number $J$.
In addition to the increased prediction difficulty, a larger $J$ requires longer training samples to provide both historical context and supervision for all forecast tokens, as quantified below.

To quantify this training-coverage constraint, we measure the proportion of PreSTS training samples whose total length satisfies $T_{\mathrm{sample}}\geq 2H_{\mathrm{step}}$, where $H_{\mathrm{step}}=JH_s$ and $H_s=64$.
This criterion identifies samples that can provide at least $H_{\mathrm{step}}$ historical observations and supervision for all $J$ future patches.
Table~\ref{tab:decoder_training_coverage} shows that this proportion decreases as $J$ increases.

As $J$ increases, fewer training samples can provide both sufficient history and supervision for all forecast tokens, potentially limiting the training of later forecast patches.
Together with the forecasting results in Figure~\ref{fig:patch_number_performance}, these statistics support choosing $J=2$ as a balance between parallel prediction and training coverage.

\begin{table}[htbp]
\centering
\caption{PreSTS training coverage as the number of forecast tokens increases, with fixed patch length $H_s=64$.}
\label{tab:decoder_training_coverage}
\begin{tabular}{ccc}
\toprule
\textbf{Forecast tokens $J$} & \textbf{Points per round $H_{\mathrm{step}}$} & \textbf{Samples with $T_{\mathrm{sample}}\geq 2H_{\mathrm{step}}$} \\
\midrule
1 & 64 & 85.73\% \\
2 & 128 & 82.69\% \\
6 & 384 & 75.54\% \\
12 & 768 & 71.53\% \\
\bottomrule
\end{tabular}
\end{table}

\minisection{Controlling the Direct Forecast Length}
To distinguish the benefit of multiple forecast tokens from that of a longer direct forecast, we compare $(J=1,H_s=128)$ with $(J=2,H_s=64)$ on GIFT-Eval.
Both settings predict 128 future time points per decoding round.
As shown in Table~\ref{tab:decoder_matched_horizon}, the two-token configuration achieves lower normalized MASE across all horizon groups.
Thus, the improvement cannot be explained solely by increasing the number of points predicted per round.
Separate learnable queries allow the decoder to retrieve information for different future patches, alleviating the need to encode the entire direct forecast in a single output embedding.

\begin{table}[htbp]
\centering
\caption{Decoder comparison at a fixed direct forecast length of 128. Results are normalized MASE on GIFT-Eval.}
\label{tab:decoder_matched_horizon}
\begin{tabular}{lcccc}
\toprule
\textbf{Setting} & \textbf{Short} & \textbf{Medium} & \textbf{Long} & \textbf{AVG} \\
\midrule
$J=1,H_s=128$ & 0.721 & 0.807 & 0.837 & 0.763 \\
$J=2,H_s=64$ & \textbf{0.709} & \textbf{0.761} & \textbf{0.794} & \textbf{0.738} \\
\bottomrule
\end{tabular}
\end{table}

\minisection{Autoregressive Forecasting Beyond One Round}
We evaluate \method-Base on ETTh1 with a context length of 2048, using $J=2$ and $H_s=64$ to predict 128 points per round.
Standard autoregressive (AR) decoding appends the median predictions to the history for subsequent rounds.
In a teacher-forced oracle, the predicted values are replaced by the corresponding ground-truth observations only in the history used by subsequent rounds; the predictions being evaluated are not replaced.
We report mean absolute error averaged over the evaluated forecast entries in each round.
The relative MAE penalty is computed as $(\mathrm{MAE}_{\mathrm{AR}}-\mathrm{MAE}_{\mathrm{oracle}})/\mathrm{MAE}_{\mathrm{oracle}}$ and reported as a percentage.

The two settings coincide in the first round because both use observed history.
Across rounds 2--5, AR decoding incurs a moderate MAE penalty of 9.79\%--14.85\% relative to the teacher-forced oracle, without a monotonic increase over successive rounds.
These results suggest that \method effectively controls error accumulation during multi-round autoregressive forecasting.

\begin{table}[htbp]
\centering
\caption{AR decoding versus a teacher-forced oracle on ETTh1 using \method-Base and a context length of 2048. Forecast ranges are relative to the initial forecast origin.}
\label{tab:decoder_ar_oracle}
\begin{tabular}{ccccc}
\toprule
\textbf{Round} & \textbf{Forecast range} & \textbf{AR MAE} & \textbf{Oracle MAE} & \textbf{MAE penalty} \\
\midrule
1 & 1--128 & 0.381 & 0.381 & 0.00\% \\
2 & 129--256 & 0.437 & 0.395 & +10.61\% \\
3 & 257--384 & 0.422 & 0.368 & +14.85\% \\
4 & 385--512 & 0.419 & 0.382 & +9.79\% \\
5 & 513--640 & 0.462 & 0.407 & +13.62\% \\
\bottomrule
\end{tabular}
\end{table}

\section{Validation of Learned Segment-Level Adaptation}
\label{app:segment_level_adaptation}

To explicitly demonstrate that the performance gains of the Mixture-of-Size Encoder stem from genuine segment-level adaptation rather than merely a generic local multi-scale inductive bias, we conduct two dedicated analyses: a structural comparison with a sequence-level multi-scale baseline, and a direct causal intervention on the learned routing decisions.
We further evaluate sensitivity to the granularity configuration to assess the robustness of the encoder design.
A high-frequency forecasting analysis tests whether its gains extend beyond improved local-level predictions.

\textbf{Advantage of Intra-Sequence Adaptation.} 
We separate the benefits of local intra-sequence adaptation from global multi-scale availability.
To achieve this, we replaced our Mixture-of-Size Encoder with the sequence-level multi-scale routing mechanism proposed in Pathformer~\citep{chen2024pathformer}. 
While Pathformer adaptively selects patch sizes for each sequence, it applies these selected granularities uniformly across the entire input, ignoring intra-sequence heterogeneity. As shown in Table~\ref{tab:causal_routing}, this replacement degrades GIFT-Eval MASE from 0.738 to 0.759, underscoring the advantage of local intra-sequence adaptation over global sequence-level multi-scale modeling.

\textbf{Causal Intervention on Routing Decisions.} 
To more directly test whether the router makes meaningful local decisions, we performed causal interventions on the routing mechanism at inference time. We kept all model weights and architectural components unchanged, but actively manipulated the router's outputs:
\begin{itemize}
    \setlength{\itemsep}{2pt}
    \setlength{\parskip}{0pt}
    \item \textit{Uniform Weights:} 
    Assigning uniform weights across all granularities within each segment, which removes adaptive selection while preserving the multi-scale structure.
    \item \textit{Shuffled Routing:} 
    Randomly shuffling the router-produced granularity weights within each segment, which disrupts the learned routing decisions.
\end{itemize}

Assigning uniform weights across all granularities degrades performance from 0.738 to 0.831, indicating that the gain cannot be explained by multi-scale modeling alone. When we further disrupt the learned routing policy by randomly shuffling the granularity weights, performance deteriorates much more sharply to 1.205. Taken together, these results provide direct intervention evidence that \method benefits not simply from a generic multi-scale gain, but from assigning granularities appropriately to each segment.

\begin{table}[htbp]
\centering
\caption{Empirical validation of segment-level adaptation on GIFT-Eval. Results report the normalized MASE across
prediction horizons and as an aggregate across all tasks.}
\label{tab:causal_routing}
\resizebox{0.7\textwidth}{!}{
\begin{tabular}{lcccc}
\toprule
\textbf{Setting} & \textbf{Short} & \textbf{Medium} & \textbf{Long} & \textbf{AVG} \\
\midrule
\rowcolor{gray!10} \textbf{\method (Full Model)} & \textbf{0.709} & \textbf{0.761} & \textbf{0.794} & \textbf{0.738} \\
\midrule
\multicolumn{5}{l}{\textit{\textbf{Structural Replacement}}} \\
\quad w/ Sequence-Level Routing (Pathformer) & 0.730 & 0.782 & 0.819 & 0.759 \\
\midrule
\multicolumn{5}{l}{\textit{\textbf{Causal Interventions at Inference}}} \\
\quad w/ Uniform Granularity Weights & 0.856 & 0.780 & 0.819 & 0.831 \\
\quad w/ Shuffled Routing Decisions & 1.154 & 1.282 & 1.271 & 1.205 \\
\bottomrule
\end{tabular}
}
\end{table}

\textbf{Sensitivity to the Granularity Configuration.}
Using \method-Base, we progressively extend the expert set toward finer patch sizes and evaluate normalized MASE on GIFT-Eval.
For each configuration, the routing budget $K$ equals the number of granularity experts $G$, while the number of null experts remains fixed at $Z=2$.
As shown in Table~\ref{tab:expert_sensitivity}, adding the patch-32 expert reduces MASE from 0.747 to 0.738, supporting the importance of finer-scale representations.
Adding a patch-16 expert yields only a further 0.001 reduction while allowing finer tokenization and longer token sequences.
We therefore use $\{32,64,128,256\}$ to balance forecasting accuracy and computational efficiency.

\begin{table}[htbp]
\centering
\caption{Sensitivity of \method-Base to the granularity configuration on GIFT-Eval.
The routing budget follows $K=G$, and all configurations use two null experts.}
\label{tab:expert_sensitivity}
\begin{tabular}{lccc}
\toprule
\textbf{Patch sizes} & \textbf{Experts $G$} & \textbf{Top-$K$} & \textbf{Normalized MASE $\downarrow$} \\
\midrule
$\{64,128,256\}$ & 3 & 3 & 0.747 \\
$\{32,64,128,256\}$ & 4 & 4 & 0.738 \\
$\{16,32,64,128,256\}$ & 5 & 5 & \textbf{0.737} \\
\bottomrule
\end{tabular}
\end{table}

\textbf{High-Frequency Forecasting.}
To assess whether adaptive tokenization improves high-frequency forecasting beyond predicting local levels, we evaluate high-pass signal error on GIFT-Eval.
We apply first differences, $\Delta y_t=y_t-y_{t-1}$ and $\Delta\hat{y}_t=\hat{y}_t-\hat{y}_{t-1}$, to the ground truth and forecast, respectively, and compute the MAE between the resulting signals for each task.
We normalize each task's high-pass MAE by that of Seasonal Na\"ive and report the geometric mean of these ratios across tasks, following the aggregation protocol used for the main GIFT-Eval results.
Table~\ref{tab:high_pass_error} compares \method with its fixed-patch variant and Sundial.
\method achieves the lowest normalized high-pass MAE in every horizon group, with an overall score of 0.750 versus 0.763 for fixed patching and 0.790 for Sundial.
The improvement over the fixed-patch variant supports the contribution of adaptive tokenization to forecasting fine-scale temporal changes, beyond improvements in local-level predictions.

\begin{table}[htbp]
\centering
\caption{Normalized high-pass MAE on GIFT-Eval using first differences.
Scores are normalized by Seasonal Na\"ive per task and aggregated with a geometric mean.
Lower values indicate better performance.}
\label{tab:high_pass_error}
\begin{tabular}{lcccc}
\toprule
\textbf{Model} & \textbf{AVG} & \textbf{Short} & \textbf{Medium} & \textbf{Long} \\
\midrule
\method & \textbf{0.750} & \textbf{0.791} & \textbf{0.699} & \textbf{0.703} \\
\method (fixed patch) & 0.763 & 0.804 & 0.712 & 0.714 \\
Sundial & 0.790 & 0.843 & 0.720 & 0.733 \\
\bottomrule
\end{tabular}
\end{table}

\section{Analysis of Inference Speed}
\label{app:infer_speed}

To provide a comprehensive view of our model's efficiency, we benchmarked the single-batch inference speed of \method against several state-of-the-art models. 
The experimental setup involved an input sequence length of 2048 and a prediction horizon of 96 on a single NVIDIA TITAN RTX GPU. 
For TTM-Advanced~\citep{ekambaram2024tiny}, its maximum supported input length of 1536 was used.

As shown in Table~\ref{tab:inference_speed}, \method-Base occupies a superior position on the efficiency-performance frontier.
Its inference time (0.061s) is on the same order of magnitude as other highly efficient models like ChronosBolt and Moirai, yet it delivers significantly better zero-shot accuracy (0.738 MASE).
Compared to high-capacity models like Toto and Chronos, \method is several orders of magnitude faster while maintaining higher forecasting precision.
The smaller \method-Mini also offers a favorable accuracy--latency trade-off, achieving considerably lower normalized MASE than TTM-Advanced (0.753 vs. 1.020) at approximately 3.3 times its inference time.

\begin{table}[htbp]
\centering
\caption{Comparison of single-batch inference speeds and zero-shot performance on GIFT-Eval. 
All models were tested with an input length of 2048 and an output length of 96, except for TTM-Advanced (*), which used its maximum input length of 1536. 
Inference times are averaged per batch on a single GPU.}
\label{tab:inference_speed}
\begin{tabular}{lcc}
\toprule
\textbf{Model} & \textbf{Inference Time (s) $\downarrow$} & \textbf{MASE $\downarrow$} \\
\midrule
TTM-Advanced* & \textbf{0.009} & 1.020 \\
Timer-XL          & 0.014 & 0.795 \\
\rowcolor{gray!10} \textbf{\method-Mini} & 0.030 & 0.753 \\
ChronosBolt-Base  & 0.055 & 0.808 \\
\rowcolor{gray!10} \textbf{\method-Base} & 0.061 & \textbf{0.738} \\
Moirai-Large      & 0.070 & 0.875 \\
Sundial-Base      & 0.099 & 0.750 \\
Time-MoE-Large    & 0.148 & 0.755 \\
TimesFM-2.0       & 0.202 & 0.758 \\
Chronos-Large     & 4.901 & 0.870 \\
Toto-Base         & 13.717 & 0.750 \\
\bottomrule
\end{tabular}
\end{table}

\section{Decoupling Architecture and Data Contributions}
\label{app:ablation_data}

In this section, we provide a dedicated experimental analysis to explicitly decouple the performance contributions of our proposed PreSTS corpus from the architectural innovations of \method.
We further assess architectural gains through parameter-matched and from-scratch comparisons.

\textbf{Ablation on Data Curation.}
We isolate the impact of our data curation components, specifically the predictability-based tier-stratified sampling and the inclusion of synthetic data. As shown in Table~\ref{tab:data_ablation}, removing the data curation components degrades the performance on the GIFT-Eval benchmark, confirming their utility in providing high-quality supervision. However, this degradation is notably smaller than the performance drop observed when removing the main architectural components. This indicates that while the PreSTS data curation strategy provides a complementary gain, the architectural design serves as the primary contributor to the overall performance of \method.

\begin{table}[htbp]
\centering
\caption{Ablation study decoupling data curation and architectural components. Models were evaluated using the normalized MASE on GIFT-Eval.}
\label{tab:data_ablation}
\resizebox{0.6\textwidth}{!}{
\begin{tabular}{lc}
\toprule
\textbf{Variant} & \textbf{MASE $\downarrow$} \\
\midrule
\multicolumn{2}{l}{\textit{\textbf{Data Curation Ablations}}} \\
\quad w/o tier-stratified sampling & 0.748 \\
\quad w/o tier-stratified sampling \& synthetic data & 0.755 \\
\midrule
\multicolumn{2}{l}{\textit{\textbf{Architectural Ablations}}} \\
\quad w/o Multi-Patch Decoder & 0.753 \\
\quad w/o Mixture-of-Size Encoder & 0.761 \\
\quad w/o DRoPE & 0.767 \\
\midrule
\rowcolor{gray!10} \textbf{\method (Full Model)} & \textbf{0.738} \\
\bottomrule
\end{tabular}%
}
\end{table}

\textbf{Matched-Data Comparison.} 
To further disentangle the architecture from the pre-training data, we perform a matched-data comparison. Specifically, we train \method on the original Chronos corpus~\citep{ansari2024chronos} and train the advanced baseline ChronosBolt on our proposed PreSTS corpus. The results are summarized in Table~\ref{tab:matched_data}.

\begin{table}[htbp]
\centering
\caption{Matched-data comparison on the GIFT-Eval benchmark separating architecture and training data contributions.}
\label{tab:matched_data}
\begin{tabular}{lccc}
\toprule
\textbf{Model} & \textbf{Params} & \textbf{Training Data} & \textbf{MASE $\downarrow$} \\
\midrule
\rowcolor{gray!10}{\method} & 53M & PreSTS & \textbf{0.738} \\
\method & 53M & Chronos corpus & 0.761 \\
ChronosBolt & 205M & PreSTS & 0.781 \\
ChronosBolt & 205M & Chronos corpus & 0.808 \\
Chronos & 709M & Chronos corpus & 0.870 \\
\bottomrule
\end{tabular}
\end{table}

These matched-data results demonstrate that the architectural contribution significantly outweighs the data-curation contribution in our setting:
\begin{itemize}
    \setlength{\itemsep}{2pt}
    \setlength{\parskip}{0pt}
    \item \method trained on the Chronos corpus (0.761) still outperforms ChronosBolt trained on PreSTS (0.781), despite using about 4$\times$ fewer parameters. This indicates that \method retains a clear advantage even without the proposed PreSTS corpus.
    \item Under the same Chronos corpus, the gap between \method and ChronosBolt is 0.047 (0.761 vs. 0.808), whereas for ChronosBolt the gain from replacing the Chronos corpus with PreSTS is 0.027 (0.808 to 0.781). This suggests that, in our experiments, architectural innovation is the primary source of improvement, while PreSTS provides an additional but smaller gain.
\end{itemize}

\textbf{Parameter-Matched Comparison.}
To determine whether the architectural gains persist under the same total parameter budget, we compare \method-Base with a parameter-matched Transformer.
The baseline uses fixed patching and standard RoPE, does not retain the Multi-Patch Decoder, and has its backbone enlarged to match the 53M total parameters of \method-Base.
Both models are trained on PreSTS with the same training steps and optimization settings and evaluated under the same protocol.
As shown in Table~\ref{tab:parameter_matched}, \method-Base achieves a normalized MASE of 0.738, compared with 0.797 for the parameter-matched baseline.
These results demonstrate that allocating capacity to the proposed architecture is more effective than enlarging a conventional Transformer backbone to the same total size.

\begin{table}[htbp]
\centering
\caption{Parameter-matched comparison on GIFT-Eval.
Both models use the same pre-training data, training steps, and optimization settings.}
\label{tab:parameter_matched}
\begin{tabular}{lcc}
\toprule
\textbf{Model} & \textbf{Params} & \textbf{Normalized MASE $\downarrow$} \\
\midrule
Parameter-matched Transformer & 53M & 0.797 \\
\textbf{\method-Base} & 53M & \textbf{0.738} \\
\bottomrule
\end{tabular}
\end{table}

\textbf{Training from Scratch.}
To evaluate the architecture independently of pre-training, we train \method-Mini from scratch on the Weather training split and compare it with full-shot PatchTST, DLinear, and TimesNet.
We select Weather for its larger size relative to the ETT datasets, reducing the risk of overfitting in this setting.
Table~\ref{tab:from_scratch} reports MSE and MAE at four forecast horizons and their averages.
Without any pre-training, \method-Mini achieves the lowest average MSE and MAE among the compared models, demonstrating that its architectural advantages extend to supervised forecasting without a pre-training corpus.
This experiment complements the matched-data and parameter-matched comparisons by removing pre-training altogether.

\begin{table}[htbp]
\centering
\caption{From-scratch forecasting on Weather.
Each entry reports MSE / MAE, and AVG is the arithmetic mean over the four forecast horizons.
The best value for each metric at each horizon is highlighted in \textbf{bold}.}
\label{tab:from_scratch}
\resizebox{0.7\textwidth}{!}{%
\begin{tabular}{lcccc}
\toprule
\textbf{Horizon} & \textbf{\method-Mini} & \textbf{PatchTST} & \textbf{DLinear} & \textbf{TimesNet} \\
\midrule
96 & 0.152 / \textbf{0.202} & \textbf{0.149} / 0.205 & 0.174 / 0.238 & 0.174 / 0.222 \\
192 & 0.197 / \textbf{0.246} & \textbf{0.195} / 0.248 & 0.214 / 0.272 & 0.229 / 0.265 \\
336 & \textbf{0.246 / 0.286} & 0.254 / 0.293 & 0.256 / 0.304 & 0.282 / 0.304 \\
720 & 0.316 / \textbf{0.335} & 0.346 / 0.354 & \textbf{0.309} / 0.346 & 0.360 / 0.352 \\
\midrule
AVG & \textbf{0.228 / 0.267} & 0.236 / 0.275 & 0.238 / 0.290 & 0.261 / 0.286 \\
\bottomrule
\end{tabular}%
}
\end{table}

\section{Implementation Details}
\label{app:imprementation_details}
\subsection{Training Details}
\label{app:training_details}
\subsubsection{Model Configurations}
We train \method in three sizes: mini (10M), small (23M) and base (53M) with the detailed model configurations are in Table \ref{tab:Kairos_configurations}. 
The base model are trained for 300,000 steps with batch sizes of 512.
We employ the AdamW optimizer, a linear decay learning rate adjustment strategy for model optimization. 
The learning rate for parameters related to DRoPE is set to 1e-5, while the learning rate for others is set to 1e-3.
Training is conducted on 4 × NVIDIA A100 GPUs using TF32 precision, which takes only 15 hours for base size.

\begin{table}[htbp]
    \centering
    \caption{Details of \method model configurations.}
    \label{tab:Kairos_configurations}
    \setlength{\tabcolsep}{2.6pt}
    \resizebox{\linewidth}{!}{
    \begin{tabular}{l *{14}{c}}
        \toprule
        & Layers & Heads & $D_h$ & $d_{\text{ff}}$ & $d_\text{expert}$ & $\{P_i\}_{i=1}^{G}$ & $K$ & $G$ & $Z$ & $\tau$ & $\eta_b$ & $\omega$ & $\theta$ & Params\\
        \midrule
        $\method_{\text{mini}}$
        & 4 & 4 & 256 & 1024 & 1408  & $\{32,64,128\}$ 
        & 3 & 3 & 2 & $\{0.55,0.1,0.05,0.15,0.15\}$ & 0.01 & 128 & $10000^{-2d/64}$ & 10M
        \\
        $\method_{\text{small}}$
        & 4 & 8 & 384 & 1536 & 1408  & $\{32,64,128\}$ 
        & 3 & 3 & 2 & $\{0.55,0.1,0.05,0.15,0.15\}$ & 0.01 & 128 & $10000^{-2d/64}$ & 23M
        \\
        $\method_{\text{base}}$
        & 6 & 8 & 512 & 2048 & 1408  & $\{32,64,128,256\}$ 
        & 4 & 4 & 2 & $\{0.45,0.1,0.1,0.05,0.15,0.15\}$ & 0.01 & 128 & $10000^{-2d/64}$ & 53M
        \\
        \bottomrule
    \end{tabular}}
\end{table}

\subsubsection{Loss Function}
\label{app:loss_function}
To better accommodate varied forecast horizons, and following the methodology of ElasTST~\citep{zhang2024elastst}, we build upon the standard quantile loss by assigning distinct weights to each timestep, such that earlier predictions are given greater importance.
The model’s parameters are optimized by minimizing the quantile loss with weight decay, formulated as:

\begin{equation}
\mathcal{L} = \frac{1}{B} \sum_{b=1}^B \sum_{t=1}^H \frac{1}{Q} \sum_{k=1}^Q w(t) L_{\alpha_k}(y_{b,t}, q_{b,t}(\alpha_k)),
\end{equation}

\begin{equation}
w(t) = \frac{1}{H}(\ln(H) - \ln(t')),\quad L_\alpha(y,q) = (\alpha - \mathbf{1}_{\{ y < q \}} )(y-q),
\end{equation}
where $B, H, Q$ are the batch size, forecast horizon, and total number of quantiles, respectively; $y_{b,t}$ is the ground truth value, $q_{b,t}(\alpha_k)$ is the $k$-th quantile forecast, $w(t)$ is the weight at each time step $t$, and $L_{\alpha_k}$ denotes the pinball loss function.
To prevent the loss weight from evaluating to zero at the final prediction step ($t=H$), we evaluate the logarithmic decay using $t'$, which maps the discrete timesteps $t \in \{1, \dots, H\}$ to an evenly spaced continuous sequence from $1 + \epsilon_1$ to $H - \epsilon_2$ (specifically, $\epsilon_1=10^{-5}$ and $\epsilon_2=10^{-3}$ in our implementation).
Empirically, we set $Q = 9$ and use quantile levels of $\alpha \in \{0.1, 0.2, \dots, 0.9\}$.
The point forecast $\hat{\mathbf{Y}}$ in the main text corresponds to the median prediction, i.e., $\hat{y}_{b,t}=q_{b,t}(0.5)$.

\subsection{Evaluation Details}

\subsubsection{Metric}
\label{app:eval_metric}
\minisection{GIFT-Eval}
GIFT-Eval employs the Mean Absolute Scaled Error (MASE) and the Continuous Ranked Probability Score (CRPS) to assess the performance of point and probabilistic forecasts, respectively.
Following the official evaluation protocol, we normalize the metrics for each task using a seasonal naïve baseline and subsequently aggregate the scores across all tasks via the geometric mean.

\minisection{TSLib} We adopt mean square error (MSE) and mean absolute error (MAE) as evaluation metrics. These metrics are calculated as follows:
\begin{equation}
\text{MSE} = \frac{1}{H}\sum_{i=1}^H (y_i - \hat{y}_i)^2, \quad \text{MAE} = \frac{1}{H}\sum_{i=1}^H \lvert y_i - \hat{y}_i\rvert
\end{equation}
where $y_i$ is the ground truth and $\hat{y}_i$ is the prediction for the $i$-th future time point.

\subsubsection{Stride}
Due to the significant inference latency of our baseline, Chronos~\citep{ansari2024chronos}, we set the evaluation stride to 96 to enhance evaluation efficiency without compromising fairness.
This setting is consistent with the protocols of Moirai~\citep{woo2024moirai}, Moirai-MoE~\citep{liu2024moirai}, and the GIFT-Eval benchmark~\citep{aksu2024gifteval}, all of which set the stride equal to the minimum prediction length.
Additionally, the original Chronos paper evaluates the model on only a single window per dataset. 
Therefore, we wish to emphasize a critical point: while the specific numerical results would likely vary with a different stride, our chosen protocol is applied uniformly to all models under evaluation, ensuring a fair and equitable comparison.

\subsection{Mixture-of-Size Encoder}
\subsubsection{Auxiliary-Loss-Free Load Balancing}
\label{app:bias}
In this section, we elaborate on the details of bias previously introduced in Section~\ref{sec:mos-encoder}.
In order to ensure that different experts are adequately trained and to control the distribution ratios of various patch sizes, we employ an Auxiliary-Loss-Free Load Balancing method similar to that used in DeepSeek-V3~\citep{liu2024deepseek}.

Specifically, we compute the empirical load $L_i$ for the $i$-th expert by summing its normalized routing weights across all $N$ segments of all $B$ sequences within a training batch:
\begin{equation}
L_i = \sum_{b=1}^B \sum_{n=1}^N \tilde{s}_{b,n,i},
\end{equation}
where $\tilde{s}_{b,n,i}$ denotes the routing weights for the $i$-th expert on the $n$-th segment of the $b$-th sequence, which directly corresponds to the definition of $\tilde{s}_i$ in Section~\ref{sec:mos-encoder}.

We define a target load distribution $\tau = (\tau_1, \dots, \tau_{G+Z})$, where $\tau_i$ is the desired proportion of the total load for expert $i$, satisfying $\sum_{j=1}^{G+Z} \tau_j = 1$. 
At the end of each training step, we update the bias term $\boldsymbol{b}_i$ using the empirical load $L_i$ and the target $\tau$:
\begin{equation}
\boldsymbol{b}_i \leftarrow \boldsymbol{b}_i + \eta_b \cdot \frac{\tau_i \cdot \sum_{j=1}^{G+Z} L_j - L_i}{\sum_{j=1}^{G+Z} L_j},
\end{equation}
where $\eta_b$ is a hyper-parameter governing the magnitude of this adjustment, referred to as the bias update speed.

This dynamic adjustment of $\boldsymbol{b}_i$ modifies the routing weights $\tilde{s}_{b,n,i}$ in subsequent batches and thereby influences Top-$K$ selections.
It encourages the normalized load $L_i / \sum_{j=1}^{G+Z} L_j$ to approach the target proportion $\tau_i$, promoting expert utilization consistent with the desired distribution over time.

\subsubsection{Definition of the Expansion Function}
\label{app:expansion_def}

In this section, we provide the formal definition for the function $\mathrm{Expand}(\cdot)$ within the Mixture-of-Size Fusion stage, as introduced in Section \ref{sec:mos-encoder}.

Consider a single time series segment $\mathbf{x}$ of length $P$. During the Top-$K$ Granularity Routing, a set of valid experts $\mathcal{I}_{\mathrm{valid}}$ is activated. Let $P_{\min}$ denote the finest granularity among these activated experts:
\begin{equation}
    P_{\min} = \min \{ P_j \mid j \in \mathcal{I}_{\mathrm{valid}} \}.
\end{equation}
Consequently, the target sequence length for alignment is defined as $M = \lceil P / P_{\min} \rceil$.

For an activated granularity level $i$ with patch size $P_i$, the encoding process yields a sequence of embeddings $\boldsymbol{E}_i = [\boldsymbol{e}_{i,1}, \dots, \boldsymbol{e}_{i,M_i}]$, where $M_i = \lceil P / P_i \rceil$. Due to the nested design of patch sizes, the ratio $R_i = P_i / P_{\min}$ is strictly a positive integer.

The function $\mathrm{Expand}(\boldsymbol{E}_i)$ maps the coarser embedding sequence to the finest temporal resolution by repeating each embedding vector $R_i$ times.
Formally, the $m$-th embedding of the expanded sequence $\tilde{\boldsymbol{E}}_i \in \mathbb{R}^{M \times D_h}$ is derived from the original sequence via index mapping:
\begin{equation}
    \tilde{\boldsymbol{e}}_{i,m} = \boldsymbol{e}_{i, \lceil m / R_i \rceil}, \quad \text{for } m = 1, \dots, M.
\end{equation}
In vector notation, this operation can be expressed as:
\begin{equation}
    \mathrm{Expand}(\boldsymbol{E}_i) = [\underbrace{\boldsymbol{e}_{i,1}, \dots, \boldsymbol{e}_{i,1}}_{R_i \text{ times}}, \dots, \underbrace{\boldsymbol{e}_{i,M_i}, \dots, \boldsymbol{e}_{i,M_i}}_{R_i \text{ times}}].
\end{equation}

We now illustrate the computational process with the following example.
Let the setup be as follows:
\begin{itemize}
    \item Segment length: $P = 128$.
    \item Set of available granularity patch sizes: $\{P_1, P_2, P_3\} = \{32, 64, 128\}$.
    \item Number of null experts: $Z=2$.
    \item Top-$K$ selection: $K=3$.
\end{itemize}
Suppose for a specific segment $\mathbf{x}$, the router activates experts corresponding to granularity levels 1 and 3, alongside one null expert (i.e., the selected set is $\{1, 3, \text{null}\}$ while $\mathcal{I}_{\mathrm{valid}} = \{1, 3\}$).
The finest activated resolution is $P_{\min} = \min(32, 128) = 32$.
The target aligned sequence length is $M = 128 / 32 = 4$.

The expansion process for each expert is computed as follows:

\begin{itemize}
    \item \textbf{For Expert 1 ($P_1 = 32$):} \\
    The number of original patches is $M_1 = 128 / 32 = 4$. The embedding sequence is $\boldsymbol{E}_1 = [\boldsymbol{e}_{1,1}, \boldsymbol{e}_{1,2}, \boldsymbol{e}_{1,3}, \boldsymbol{e}_{1,4}]$.
    The repetition factor is $R_1 = 32 / 32 = 1$.
    \begin{equation}
        \tilde{\boldsymbol{E}}_1 = \mathrm{Expand}(\boldsymbol{E}_1) = [\boldsymbol{e}_{1,1}, \boldsymbol{e}_{1,2}, \boldsymbol{e}_{1,3}, \boldsymbol{e}_{1,4}].
    \end{equation}
    Since this expert operates at the finest activated resolution, the expansion is an identity mapping.

    \item \textbf{For Expert 3 ($P_3 = 128$):} \\
    The number of original patches is $M_3 = 128 / 128 = 1$. The embedding sequence consists of a single vector $\boldsymbol{E}_3 = [\boldsymbol{e}_{3,1}]$.
    The repetition factor is $R_3 = 128 / 32 = 4$.
    \begin{equation}
        \tilde{\boldsymbol{E}}_3 = \mathrm{Expand}(\boldsymbol{E}_3) = [\boldsymbol{e}_{3,1}, \boldsymbol{e}_{3,1}, \boldsymbol{e}_{3,1}, \boldsymbol{e}_{3,1}].
    \end{equation}
    The single coarse embedding is broadcast across all four aligned token positions to match the target resolution.

    \item \textbf{Hypothetical Case for Expert 2 ($P_2 = 64$):} \\
    Although Expert 2 was not activated in this example, if it were processed, $M_2 = 128 / 64 = 2$ and $R_2 = 64 / 32 = 2$. The expansion would be:
    \begin{equation}
        \tilde{\boldsymbol{E}}_2 = [\boldsymbol{e}_{2,1}, \boldsymbol{e}_{2,1}, \boldsymbol{e}_{2,2}, \boldsymbol{e}_{2,2}].
    \end{equation}
\end{itemize}

By explicitly aligning all representations to the finest resolution $P_{\min}$ via $\mathrm{Expand}(\cdot)$, \method ensures that the subsequent weighted summation $\bar{\boldsymbol{E}} = \sum_{i \in \mathcal{I}_{\mathrm{valid}}} c_i \, \tilde{\boldsymbol{E}}_i$ is performed on element-wise compatible tensors, effectively fusing global context with local details.

\subsection{Dynamic Rotary Position Embedding (DRoPE)}
\label{app:DRoPE}
In this section, we provide a more detailed description of the DRoPE implementation discussed in Section~\ref{method:DRoPE}. 
We first introduce the principles of Rotary Position Embedding (RoPE)~\citep{su2024roformer} in detail, and then provide corresponding supplementary information on the DRoPE implementation.

\subsubsection{Details of RoPE}
\label{app:detail_of_rope}
The core idea of RoPE is to rotate segments of the query and key vectors by an angle proportional according to their absolute position in the sequence. This allows the model to discern relative positions through the geometry of these rotations, without needing explicit calculation of relative distances.
Throughout this subsection, we use $m$ and $n$ to denote generic token positions.

Specifically, RoPE operates on vectors of an even rotation dimension, denoted as $D_{\mathrm{rot}}$.
For a vector $\boldsymbol{e}$ (representing a query $\mathbf{q}$ or a key $\mathbf{k}$ in self-attention) at position $m$, it is transformed by a rotation matrix $\mathbf{R}_m$.
This matrix is block-diagonal, composed of $D_{\mathrm{rot}}/2$ individual $2 \times 2$ rotation blocks.
Each block $\mathbf{R}_{m,d}$ acts on a pair of dimensions $(e_{2d}, e_{2d+1})$ of the vector:

As introduced in Section~\ref{method:DRoPE}, RoPE applies a rotation to input vector.
This rotation is applied to pairs of dimensions $(e_{2d}, e_{2d+1})$, and can be concisely expressed using complex form:

\begin{equation}
f_{\text{RoPE}}(\boldsymbol{z}, m) = (z_{2d} + i z_{2d+1})e^{im\theta_d},
\end{equation}

where $\theta_d$ is the angular frequency.
Here, $i$ denotes the imaginary unit.
This multiplication by $e^{im\theta_d}$ corresponds to a rotation in the complex plane.
For the real-valued components $z_{2d}$ and $z_{2d+1}$, this operation is equivalent to applying the following $2 \times 2$ rotation matrix $\mathbf{R}_{m,d}$:

\begin{equation}
\mathbf{R}_{m,d} = \begin{bmatrix}
\cos(m\theta_d) & -\sin(m\theta_d) \\
\sin(m\theta_d) & \cos(m\theta_d)
\end{bmatrix},
\end{equation}

Here, $m$ is the absolute position of the token.
The term $\theta_d$ represents a predefined angular frequency for the $d$-th pair of dimensions ($d \in [0, D_{\mathrm{rot}}/2-1]$), typically defined as $\theta_d = b^{-2d/D_{\mathrm{rot}}}$.

The full rotation matrix $\mathbf{R}_m$ for position $m$ is thus:
\begin{align}
\mathbf{R}_m &= \text{diag}(\mathbf{R}_{m,0}, \mathbf{R}_{m,1}, \dots, \mathbf{R}_{m,D_{\mathrm{rot}}/2-1}) \\
             &= \begin{pmatrix}
\cos m\theta_0 & -\sin m\theta_0 & 0 & 0 & \cdots & 0 & 0 \\
\sin m\theta_0 & \cos m\theta_0 & 0 & 0 & \cdots & 0 & 0 \\
0 & 0 & \cos m\theta_1 & -\sin m\theta_1 & \cdots & 0 & 0 \\
0 & 0 & \sin m\theta_1 & \cos m\theta_1 & \cdots & 0 & 0 \\
\vdots & \vdots & \vdots & \vdots & \ddots & \vdots & \vdots \\
0 & 0 & 0 & 0 & \cdots & \cos m\theta_{D_{\mathrm{rot}}/2-1} & -\sin m\theta_{D_{\mathrm{rot}}/2-1} \\
0 & 0 & 0 & 0 & \cdots & \sin m\theta_{D_{\mathrm{rot}}/2-1} & \cos m\theta_{D_{\mathrm{rot}}/2-1}
\end{pmatrix}.
\end{align}

Let $\mathbf{q}_m$ and $\mathbf{k}_n$ be the original query and key vectors for tokens at positions $m$ and $n$ respectively. After applying RoPE, their new representations become 
$\mathbf{q}'_m = \mathbf{R}_m \mathbf{q}_m$ and $\mathbf{k}'_n = \mathbf{R}_n \mathbf{k}_n$. 
The dot product for attention is then:
\begin{equation}
(\mathbf{q}'_m)^\top (\mathbf{k}'_n) = (\mathbf{R}_m \mathbf{q}_m)^\top (\mathbf{R}_n \mathbf{k}_n) = \mathbf{q}_m^\top \mathbf{R}_m^\top \mathbf{R}_n \mathbf{k}_n, 
\end{equation}
due to the properties of rotation matrices, $\mathbf{R}_m^\top \mathbf{R}_n = \mathbf{R}_{n-m}$ we can get:
\begin{equation}
\label{eq:rope}
(\mathbf{q}'_m)^\top (\mathbf{k}'_n) = \mathbf{q}_m^\top \mathbf{R}_{n-m} \mathbf{k}_n. 
\end{equation}
This equation shows that the dot product between a query and key vector after rotation inherently depends on their original values ($\mathbf{q}_m, \mathbf{k}_n$) and their relative positions ($n-m$). This allows RoPE to integrate relative positional information into the self-attention without any additional learnable parameters or explicit relative position computations.

Equation~\ref{eq:rope} confirms that the interaction relies on the relative position $n-m$ through the full rotation matrix $\mathbf{R}_{n-m}$. Since this matrix is block-diagonal, the total interaction is actually composed of independent rotations on disjoint 2D subspaces. The following theorem demonstrates the periodicity of the attention score within each subspace.
\begin{theorem}[Periodic dependence of RoPE attention]
\label{thm:rope-periodic}
Consider the $d$-th 2D subspace of the hidden state under RoPE, with angular
frequency $\theta_d$, and let $\mathrm{atten}_d(m,n)$ denote this subspace's
contribution to the dot-product attention between positions $m$ and $n$.
Then there exist an amplitude $A_d(m,n) \ge 0$ and a phase
$\varphi_d(m,n)$, depending only on the content vectors at $m$ and $n$, such
that
\begin{equation}
\begin{split}
\mathrm{atten}_d(m,n)
  &= A_d(m,n)\,\cos\big(\theta_d (m-n) + \varphi_d(m,n)\big) \\
  &\;\propto\; \cos\big(\theta_d (m-n) + \varphi_d(m,n)\big).    
\end{split}
\end{equation}

In particular, for fixed content, $\mathrm{atten}_d(m,n)$ is a periodic
function of the relative distance $(m-n)$ with period $2\pi / \theta_d$.
\end{theorem}

\begin{proof}
Let $q_{m,d}, k_{n,d} \in \mathbb{C}$ be the complex-valued representations of
the $d$-th 2D subspace for the query and key at positions $m$ and $n$,
respectively, as defined above. Applying RoPE multiplies these components by
phase factors $e^{i m \theta_d}$ and $e^{i n \theta_d}$, yielding
\begin{equation}
q'_{m,d} = q_{m,d} e^{i m \theta_d}, \qquad
k'_{n,d} = k_{n,d} e^{i n \theta_d}.
\end{equation}
The contribution of this subspace to the dot-product attention is the real
part of $q'_{m,d} (k'_{n,d})^*$:
\begin{equation}
\mathrm{atten}_d(m,n)
  = \mathrm{Re}\!\big[q'_{m,d} (k'_{n,d})^*\big] 
  = \mathrm{Re}\!\big[q_{m,d} k_{n,d}^* e^{i \theta_d (m-n)}\big].    
\end{equation}
Writing $q_{m,d} k_{n,d}^*$ in polar form as
$q_{m,d} k_{n,d}^* = A_d(m,n) e^{i \varphi_d(m,n)}$ with $A_d(m,n) \ge 0$ and
$\varphi_d(m,n) \in \mathbb{R}$, we obtain
\begin{equation}
\begin{split}
\mathrm{atten}_d(m,n)
  &= \mathrm{Re}\!\big[A_d(m,n) e^{i (\theta_d (m-n) + \varphi_d(m,n))}\big] \\
  &= A_d(m,n)\,\cos\big(\theta_d (m-n) + \varphi_d(m,n)\big),
\end{split}
\end{equation}
which shows that $\mathrm{atten}_d(m,n)$ is proportional to a cosine function
of the relative distance $(m-n)$, with angular frequency $\theta_d$.
The periodicity with period $2\pi/\theta_d$ follows immediately from the
periodicity of the cosine function.
\end{proof}

\subsubsection{Details of DRoPE}
\label{app:drope_details}
The initial RoPE frequencies $\theta_d$ typically exhibit a wide numerical range.
For instance, in our setting $\theta_0=1.0\ (d=0)$, while $\theta_{31} \approx 1.33 \times 10^{-4}$ ($d=31$, corresponding to $D_{\mathrm{rot}}/2-1$ and $D_{\mathrm{rot}}=64$).
Directly applying affine modulation to these values could lead to numerical instability or disproportionate adjustments due to the vast scale differences.

To address this and ensure stable and effective modulation across the entire range of base frequencies, our  DRoPE performs the adaptation in log-space. 
The layer-specific modulation parameters, $\gamma^{(l)}_d$ and $\beta^{(l)}_d$, predicted by the Multilayer Perceptron (MLP) for layer $l$ following Algorithm~\ref{alg:gen_params}, are applied to the log-transformed base frequencies $\log \theta_d$ via an element-wise affine transformation as
\begin{equation}
\log\theta^{(l)}_{\text{inst},d} = \gamma^{(l)}_d \cdot \log \theta_d + \beta^{(l)}_d, 
\end{equation} 
where $\log\theta^{(l)}_{\text{inst},d}$ represents the modulated log-frequencies for layer $l$ and dimension pair $d$.
Then, these modulated log-frequencies are transformed back to their original scale by exponentiation to obtain the adaptive rotation frequencies $\theta^{(l)}_{\text{inst},d}$ used in the DRoPE calculation for layer $l$ as
\begin{equation}
\theta^{(l)}_{\text{inst},d} = \exp(\log\theta^{(l)}_{\text{inst},d}).
\end{equation}

\begin{algorithm}[H]
\caption{Generating Instance Adaptive Parameters ($\gamma^{(l)}, \beta^{(l)}$)}
\label{alg:gen_params}
\begin{algorithmic}[1]
\REQUIRE
    Time series instance $X \in \mathbb{R}^{B \times L}$, mask $M \in \{0,1\}^{B \times L}$, batch size $B$, sequence length $L$, FFT feature dimension $\omega$, MLP network $f_{\text{MLP}}$
\ENSURE
    Adaptive parameters $\gamma^{(l)} \in \mathbb{R}^{B \times D_{\mathrm{rot}}/2}$, $\beta^{(l)} \in \mathbb{R}^{B \times D_{\mathrm{rot}}/2}$

\vspace{0.5em}

\STATE $X_{\text{masked}} \gets X \odot M$ \COMMENT{Apply mask (element-wise product)}
\STATE $F_{\text{result}} \gets \text{RFFT}(X_{\text{masked}}), F_{\text{result}} \in \mathbb{C}^{B \times (\frac{L}{2}+1)}$ \COMMENT{Real FFT along sequence dim}
\STATE $F_{\text{amp}} \gets |F_{\text{result}}|, F_{\text{amp}} \in \mathbb{R}^{B \times (\frac{L}{2}+1)}$ \COMMENT{Amplitude spectrum}

\STATE $X_{\text{FFT}} \gets F_{\text{amp}}[\dots, :\omega], X_{\text{FFT}} \in \mathbb{R}^{B \times \omega}$ \COMMENT{Extract low frequency}    

\STATE $X'_{\text{FFT}} \gets \text{LayerNorm}_{\text{FFT}}(X_{\text{FFT}})$ \COMMENT{Normalize FFT features}
\STATE $\gamma^{(l)}, \beta^{(l)} \gets f_{\text{MLP}}(X'_{\text{FFT}})$ \COMMENT{Get instance-specific parameters}
\STATE \textbf{return} $\gamma^{(l)}, \beta^{(l)}$

\end{algorithmic}
\end{algorithm}

\subsubsection{Interpretation of DRoPE}
From the theorem~\ref{thm:rope-periodic}, the base RoPE parameters $\{\theta_d\}_{d=0}^{D_{\mathrm{rot}}/2-1}$ define a fixed bank of sinusoidal kernels over relative positions.
Each $\theta_d$ controls the frequency of one cosine kernel. 
This fixed frequencies works reasonably well when sequences share a homogeneous notion of temporal scale, but it can become suboptimal for a TSFM that must handle highly heterogeneous sampling intervals and spectral patterns across domains.

DRoPE addresses this by making these frequencies instance-adaptive. 
The FFT-based module in Algorithm~\ref{alg:gen_params} extracts simple spectral statistics $X'_{\text{FFT}}$ from each input sequence and maps them, via an MLP, to instance- and layer-specific parameters $\gamma^{(l)}_d$ and $\beta^{(l)}_d$. 
These parameters modulate the base frequencies in log-space:
\begin{equation}
\log \theta^{(l)}_{\text{inst},d} = \gamma^{(l)}_d \cdot \log \theta_d + \beta^{(l)}_d, 
\quad
\theta^{(l)}_{\text{inst},d} = \exp\big(\log \theta^{(l)}_{\text{inst},d}\big),
\end{equation}
yielding an adapted frequency grid $\{\theta^{(l)}_{\text{inst},d}\}$ for each layer and sequence.

Intuitively, this log-space affine transformation stretches or compresses the original RoPE frequency grid in a sequence-dependent way, while preserving the relative-position nature of RoPE. 
The resulting $\theta^{(l)}_{\text{inst},d}$ still define sinusoidal kernels over relative lags, but their effective frequencies are now gently steered by the spectrum of the current input. 
Crucially, DRoPE does not aim to recover a single true period for each series; real-world time series are often multi-periodic and non-stationary. 
Instead, we interpret $\{\theta^{(l)}_{\text{inst},d}\}$ as a sequence-dependent frequency profile that shapes how attention depends on relative lags, providing a more flexible and data-driven positional bias than fixed RoPE.

\section{Additional Details of Experiment Setting}
\label{app:additional_details}
\subsection{Pre-training Datasets}
\label{app:data}
We trained \method on the Predictability-Stratified Time Series (PreSTS) corpus, which consists of over 300B real-world time series observations from Chronos~\citep{ansari2024chronos} and Moirai~\citep{woo2024moirai} in conjunction with 15B synthetic time points.
Following~\citep{das2024decoder}, the training loader samples 80\% real data and 20\% synthetic data.

\minisection{Real-world data}
The real-world datasets were stratified into five tiers based on their predictability.
This hierarchical structure dictates the sampling probability during model training, assigning a higher likelihood of selection to datasets with greater predictability.
Such a strategy ensures that the model is preferentially trained on high-quality data while preserving its capacity to predict corner cases. 
Specifically, Tier 1 comprises datasets characterized by pronounced periodicity and trends with low noise.
Tier 2 contains datasets with similarly distinct patterns but high noise, whereas Tier 3 includes those with subtle trends and considerable noise.
The remaining datasets were classified into Tiers 4 and 5, based on a composite assessment of their size and pattern regularity.
The specific details of these datasets, categorized by their respective sampling frequencies, are presented in Tables \ref{tab:hourly_dataset}-\ref{tab:monthly_dataset}.

\minisection{Synthetic data}
We build a synthetic data generator that produces two distinct types of time series, each with a length of 4096. 
The first type consists of composite series, created by the additive combination of seasonal, trend, and noise components. 
For seasonality, we sample one or two components, where the primary period is drawn from \{24, 48, 288, 360\}, and a potential second period is a fixed seven-fold multiple of the first;
the seasonal patterns manifest as either spike trains or smooth non-sinusoidal templates generated via interpolation.
An optional trend is chosen from linear, exponential, or an ARIMA-like process derived from cumulating a stationary ARMA model's output.
High-probability white Gaussian noise is also added, with each series guaranteed to contain at least one seasonal or trend component and all magnitudes bounded for stability.
The generation process for these composite series is formally detailed in Algorithm \ref{alg:composite_gen}.
Complementing these are idealized industrial signals, which simulate perfectly regular machine cycles.
These feature a constant baseline with repeating events like trapezoidal spikes or inverted-U shaped dips, where the period, amplitude, and width of the events remain fixed across the entire series with no random jitter, as outlined in Algorithm \ref{alg:industrial_gen}.
Synthetic dataset cases are provided in Appendix \ref{app:synthetic_cases}.

\begin{figure}[htbp]
    \centering
    \includegraphics[width=0.98\textwidth]{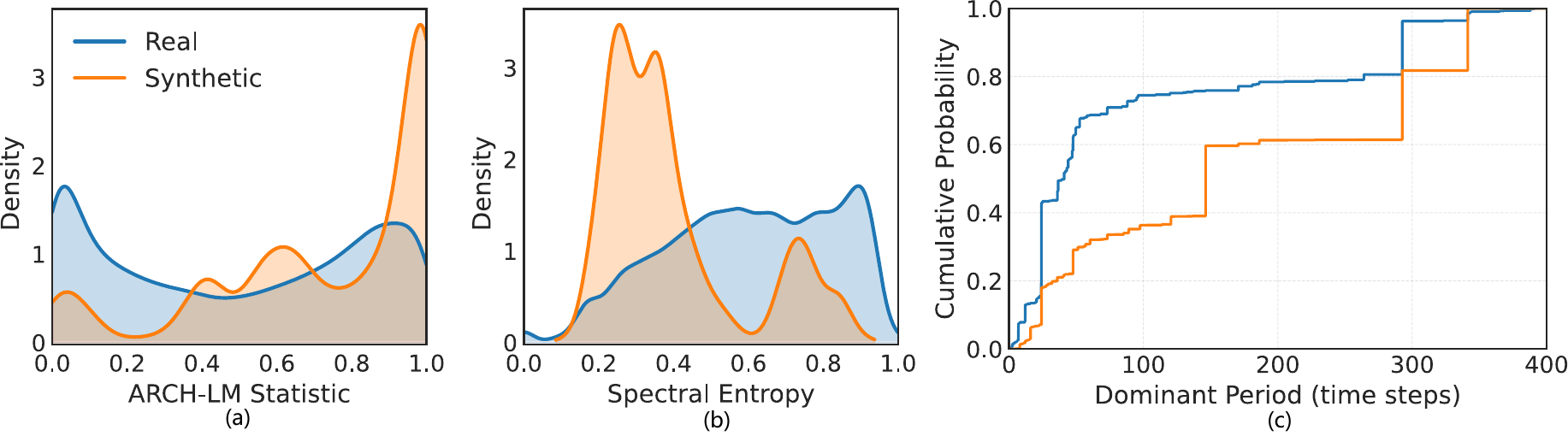}
    \caption{Comparison of real and synthetic datasets across \textbf{(a)} ARCH-LM statistic, \textbf{(b)} spectral entropy, and \textbf{(c)} dominant period distributions, illustrating complementary volatility, complexity, and periodicity characteristics.}
    \label{fig:synthetic_distribution}
\end{figure}

\minisection{Distribution Analysis}
To investigate whether the synthetic dataset supplements distributions absent in the real-world data, we adopted the statistical methodology of Toto~\citep{cohen2025time}. We employed the ARCH-LM Statistic and Spectral Entropy to quantify time-varying volatility and information density (complexity), respectively. These metrics align closely with the design motivation for the Mixture-of-Size Encoding and DRoPE components of \method.
As illustrated in Figures~\ref{fig:synthetic_distribution} (a) and (b), we observed that a significant portion of the real-world dataset exhibits ARCH-LM Statistic values approaching zero and Spectral Entropy values nearing one, indicating high time-varying volatility and complexity. In contrast, the synthetic dataset exhibits the opposite characteristics, exposing the model during training to time series with more constant volatility and greater regularity. This exposure enhances the model's generalization capabilities. 
Furthermore, we statistically compared the periodicity of the real-world and synthetic datasets by plotting their respective Cumulative Distribution Functions (CDFs).
As shown in Figure \ref{fig:synthetic_distribution} (c), the dominant periods in the real-world dataset are heavily concentrated.
This can cause the model to overfit to these specific periodicities and fail to learn generalizable periodic patterns.
The synthetic dataset complements this distribution with training examples covering different period ranges.
Consequently, the synthetic dataset effectively mitigates these biases present in the real-world data, thereby exposing the model to a more comprehensive and diverse set of distributions.

\begin{table}[htbp]
    \centering
    \caption{Detailed descriptions of second-level, minute-level, and hourly datasets.}
    \label{tab:hourly_dataset}
    \resizebox{0.7\linewidth}{!}{
    \begin{tabular}{l *{4}{c}}
        \toprule
        Dataset & Domain & Frequency & \# Time Series & \# Time points\\
        \midrule
        Wind Power & Energy & 4S & 1 & 7,397,147 \\
        Residential Load Power & Energy & T & 813 & 437,983,677 \\
        Residential PV Power & Energy & T & 699 & 376,016,850 \\
        Los-Loop & Transport & 5T & 207 & 7,094,304 \\
        PEMS03 & Transport & 5T & 358 & 9,382,464 \\
        PEMS04 & Transport & 5T & 921 & 15,649,632 \\
        PEMS07 & Transport & 5T & 883 & 24,921,792 \\
        PEMS08 & Transport & 5T & 510 & 9,106,560 \\
        PEMS Bay & Transport & 5T & 325 & 16,941,600 \\
        Alibaba Cluster Trace 2018 & CloudOps & 5T & 116,818 & 190,385,060\\
        Azure VM Traces 2017 & CloudOps & 5T & 159,472 & 885,522,908 \\
        Borg Cluster Data 2011 & CloudOps & 5T & 286,772 & 1,075,105,708 \\
        LargeST & Transport & 5T & 42,333 & 4,452,510,528 \\
        KDD Cup 2022 & Energy & 10T & 134 & 4,727,519 \\
        HZMetro & Transport & 15T & 160 & 380,320 \\
        Q-Traffic & Transport & 15T & 45,148 & 264,386,688 \\
        SHMetro & Transport & 15T & 576 & 5,073,984 \\
        Beijing Subway & Transport & 30T & 552 & 867,744 \\
        Elecdemand & Energy & 30T & 1 & 17,520 \\
        Australian Electricity Demand & Energy & 30T & 5 & 1,155,264 \\
        London Smart Meters & Energy & 30T & 5,560 & 166,528,896 \\
        Taxi & Transport & 30T & 2428 & 3,589,798 \\
        BDG-2 Bear & Energy & H & 91 & 1,482,312 \\
        BDG-2 Fox & Energy & H & 135 & 2,324,568 \\
        BDG-2 Panther & Energy & H & 105 & 919,800 \\
        BDG-2 Rat & Energy & H & 280 & 4,728,288 \\
        Borealis & Energy & H & 15 & 83,269 \\
        BDG-2 Bull & Energy & H & 41 & 719,304 \\
        China Air Quality & Nature & H & 2,622 & 34,435,404 \\
        BDG-2 Cockatoo & Energy & H & 1 & 17,544 \\
        Covid19 Energy & Energy & H & 1 & 31,912 \\
        ELF & Energy & H & 1 & 21,792 \\
        GEF12 & Energy & H & 20 & 788,280 \\
        GEF14 & Energy & H & 1 & 17,520 \\
        GEF17 & Energy & H & 8 & 140,352 \\
        BDG-2 Hog & Energy & H & 24 & 421,056 \\
        IDEAL & Energy & H & 217 & 1,255,253 \\
        Low Carbon London & Energy & H & 713 & 9,543,553 \\
        Oikolab Weather & Climate & H & 8 & 800,456 \\
        PDB & Energy & H & 1 & 17,520 \\
        Sceaux & Energy & H & 1 & 34,223 \\
        SMART & Energy & H & 5 & 95,709 \\
        Spanish Energy and Weather & Energy & H & 1 & 35,064 \\
        ERCOT Load & Energy & H & 8 & 1,238,976 \\
        Mexico City Bikes & Transport & H & 494 & 38,687,004 \\
        Beijing Air Quality & Nature & H & 132 & 4,628,448 \\
        Pedestrian Counts & Transport & H & 66 & 3,132,346 \\
        Rideshare & Transport & H & 2,304 & 859,392 \\
        Traffic & Transport & H & 862 & 15,122,928 \\
        Taxi (Hourly) & Transport & H & 2,428 & 1,794,292 \\
        Uber TLC (Hourly) & Transport & H & 262 & 1,138,128 \\
        Wind Farms (Hourly) & Energy & H & 337 & 2,869,414 \\
        Weatherbench (Hourly) & Nature & H & 225,280 & 78,992,150,528
        \\
        Buildings900K & Energy & H & 1,795,256 & 15,728,237,816
        \\
        ERA5 & Climate & H & 11,059,200 & 96,613,171,200
        \\
        CMIP6 & Climate & 6H & 14,327,808 & 104,592,998,400
        \\
        \bottomrule
    \end{tabular}}
\end{table}

\begin{table}[htbp]
    \centering
    \caption{Detailed descriptions of daily datasets.}
    \label{tab:daily_dataset}
    \resizebox{0.7\linewidth}{!}{
    \begin{tabular}{l *{4}{c}}
        \toprule
        Dataset & Domain & Frequency & \# Time Series & \# Time points\\
        \midrule
        Bitcoin & Econ/Fin & D & 18 & 81,918 \\
        Covid Mobility & Transport & D & 362 & 148,602 \\
        Extended Web Traffic & Web & D & 145,063 & 370,926,091 \\
        Favorita Sales & Sales & D & 111,840 & 139,179,538 \\
        Favorita Transactions & Sales & D & 54 & 84,408 \\
        Subseasonal & Climate & D & 3,448 & 56,788,560 \\
        Subseasonal Precipitation & Climate & D & 862 & 9,760,426 \\
        Sunspot & Nature & D & 1 & 73,894 \\
        Vehicle Trips & Transport & D & 329 & 32,512 \\
        Wiki-Rolling & Web & D & 47,675 & 40,619,100 \\
        Dominick & Retail & D & 100,014 & 29,652,492 \\
        M5 & Sales & D & 30,490 & 47,649,940 \\
        Monash Weather & Climate & D & 3,010 & 43,032,000 \\
        NN5 Daily & Econ/Fin & D & 111 & 87,801 \\
        Uber TLC Daily & Transport & D & 262 & 47,422 \\
        Weatherbench (Daily) & Nature & D & 225,280 & 3,291,336,704 \\
        Wiki Daily (100k) & Web & D & 100,000 & 274,100,000 \\
        Wind Farms (Daily) & Energy & D & 337 & 119,549 \\
        Exchange Rate & Finance & D & 8 & 60,704 \\
        \bottomrule
    \end{tabular}}
\end{table}

\begin{table}[htbp]
    \centering
    \caption{Detailed descriptions of weekly datasets.}
    \label{tab:weekly_dataset}
    \resizebox{0.7\linewidth}{!}{
    \begin{tabular}{l *{4}{c}}
        \toprule
        Dataset & Domain & Frequency & \# Time Series & \# Time points\\
        \midrule
        CDC Fluview ILINet & Healthcare & W & 375 & 319,515 \\
        CDC Fluview WHO NREVSS & Healthcare & W & 296 & 167,040 \\
        Kaggle Web Traffic Weekly & Web & W & 145,063 & 16,537,182 \\
        Project Tycho & Healthcare & W & 1,258 & 1,377,707 \\
        Traffic Weekly & Transport & W & 862 & 82,752 \\
        NN5 Weekly & Econ/Fin & W & 111 & 12,543 \\
        Weatherbench (Weekly) & Nature & W & 225,280 & 470,159,360 \\
        \bottomrule
    \end{tabular}}
\end{table}

\begin{table}[htbp]
    \centering
    \caption{Detailed descriptions of monthly, quarterly, and yearly datasets.}
    \label{tab:monthly_dataset}
    \resizebox{0.7\linewidth}{!}{
    \begin{tabular}{l *{4}{c}}
        \toprule
        Dataset & Domain & Frequency & \# Time Series & \# Time points\\
        \midrule
        GoDaddy & Econ/Fin & M & 6,270 & 257,070 \\
        CIF 2016 & Econ/Fin & M & 72 & 7,108 \\
        FRED MD & Econ/Fin & M & 107 & 77,896 \\
        M1 Monthly & Econ/Fin & M & 617 & 55,998 \\
        M3 Monthly & Econ/Fin & M & 1,428 & 167,562 \\
        Tourism Monthly & Econ/Fin & M & 366 & 109,280 \\
        M3 Other & Econ/Fin & Q & 174 & 11,933 \\
        M1 Quarterly & Econ/Fin & Q & 203 & 9,944 \\
        M3 Quarterly & Econ/Fin & Q & 756 & 37,004 \\
        Tourism Quarterly & Econ/Fin & Q & 427 & 42,544 \\
        M1 Yearly & Econ/Fin & Y & 181 & 4,515 \\
        M3 Yearly & Econ/Fin & Y & 645 & 18,319 \\
        Tourism Yearly & Econ/Fin & Y & 518 & 12,757 \\
        \bottomrule
    \end{tabular}}
\end{table}

\begin{algorithm}[htbp]
\caption{Composite Time Series Generation}
\label{alg:composite_gen}
\begin{algorithmic}[1]
\REQUIRE 
    Time series length $L=4096$, 
    Primary period set $\mathcal{P}_1 = \{24, 48, 288, 360\}$, 
    Harmonic multiplier $n=7$, 
    Trend types $\mathcal{T}_{\text{types}} = \{\text{linear, exp, ARMA}\}$, 
    Seasonal patterns $\mathcal{S}_{\text{patterns}} = \{\text{spike, interpolated segment}\}$.
\ENSURE 
    A synthetic time series $\boldsymbol{x}_{1:L}$.
\STATE $\boldsymbol{x}, \boldsymbol{s}, \boldsymbol{t}, \boldsymbol{n} \leftarrow \boldsymbol{0}_{1:L}$ \COMMENT{Initialize total series and components (seasonal, trend, noise)}
\STATE Sample flags $f_s, f_t, f_n$ \COMMENT{Determine component inclusion, ensuring $f_s \lor f_t$ is true}
\IF{$f_s$ is true}
    \STATE $k \sim \text{Bernoulli}(0.2)$ \COMMENT{$k=1$ for double period (20\% prob), $k=0$ for single}
    \STATE $p_1 \sim \mathcal{U}(\mathcal{P}_1)$ \COMMENT{Sample primary period}
    \STATE $\mathcal{P}_{\text{active}} \leftarrow \{p_1\}$
    \IF{$k=1$}
        \STATE $\mathcal{P}_{\text{active}} \leftarrow \mathcal{P}_{\text{active}} \cup \{n \cdot p_1\}$
    \ENDIF
    \FORALL{$p$ in $\mathcal{P}_{\text{active}}$}
        \STATE $a \sim \mathcal{U}(1.0, 3.0)$ \COMMENT{Sample amplitude for this component}
        \STATE $pattern \sim \mathcal{U}(\mathcal{S}_{\text{patterns}})$
        \STATE $\boldsymbol{c} \leftarrow \text{GeneratePattern}(pattern, p, a)$ \COMMENT{Create a single cycle of the pattern}
        \STATE $\boldsymbol{s} \leftarrow \boldsymbol{s} + \text{Tile}(\boldsymbol{c}, L)$ \COMMENT{Tile the cycle to length $L$ and add to seasonal component}
    \ENDFOR
\ENDIF
\IF{$f_t$ is true}
    \STATE $type \sim \mathcal{U}(\mathcal{T}_{\text{types}})$
    \STATE $\boldsymbol{t} \leftarrow \text{GenerateTrend}(type, L)$
    \IF{$f_s$ is true}
        \STATE $\lambda \sim \mathcal{U}(0.1, 0.3)$ \COMMENT{Reduce trend strength when seasonality is present}
        \STATE $\boldsymbol{t} \leftarrow \lambda \cdot \boldsymbol{t}$
    \ENDIF
\ENDIF
\IF{$f_n$ is true}
    \STATE $\sigma_n \sim \mathcal{U}(0.01, 0.1)$
    \STATE $\boldsymbol{n} \sim \mathcal{N}(\boldsymbol{0}, \sigma_n^2 \mathbf{I})$ \COMMENT{Generate white Gaussian noise}
\ENDIF
\STATE $\boldsymbol{x} \leftarrow \boldsymbol{s} + \boldsymbol{t} + \boldsymbol{n}$
\STATE \textbf{return} $\boldsymbol{x}_{1:L}$
\end{algorithmic}
\end{algorithm}

\begin{algorithm}[htbp]
\caption{Idealized Industrial Signal Generation}
\label{alg:industrial_gen}
\begin{algorithmic}[1]
\REQUIRE 
    Time series length $L=4096$, Pattern types $\mathcal{P}_{\text{types}} = \{\text{inverted\_u, spikes}\}$.
\ENSURE 
    A synthetic time series $\boldsymbol{x}_{1:L}$.
\STATE $type \sim \mathcal{U}(\mathcal{P}_{\text{types}})$
\STATE $(b, p, a, w, \sigma_n) \leftarrow \text{SampleParams}(type)$ \COMMENT{Sample baseline, period, amplitude, width, noise}
\STATE $\boldsymbol{x}_{1:L} \leftarrow b$ \COMMENT{Initialize series with baseline value}
\STATE $\boldsymbol{e} \leftarrow \text{TrapezoidShape}(w, a)$ \COMMENT{Create the event shape of width $w$ and amplitude $a$}
\IF{$type = \text{inverted\_u}$}
    \STATE $sign \leftarrow -1$
\ELSE
    \STATE $sign \leftarrow +1$
\ENDIF
\FORALL{$i \leftarrow 0, p, 2p, \dots$ up to $L-1$}
    \STATE $start \leftarrow i$, $end \leftarrow \min(i+w, L)$
    \STATE $\boldsymbol{x}_{start:end} \leftarrow \boldsymbol{x}_{start:end} + sign \cdot \boldsymbol{e}_{1:end-start}$ \COMMENT{Add or subtract event shape at periodic intervals}
\ENDFOR
\IF{$\sigma_n > 0$}
    \STATE $\boldsymbol{x} \leftarrow \boldsymbol{x} + \mathcal{N}(\boldsymbol{0}, \sigma_n^2 \mathbf{I})$ \COMMENT{Add global Gaussian noise}
\ENDIF
\STATE \textbf{return} $\boldsymbol{x}_{1:L}$
\end{algorithmic}
\end{algorithm}

\subsection{Compared Methods}
\label{app:compared_methods}
We compare \method with several state-of-the-art models, including TSFMs such as Sundial~\citep{liu2025sundial}, Toto~\citep{cohen2025time}, YingLong~\citep{wang2025output}, VisionTS~\citep{chenvisionts}, Time-MoE~\citep{shi2024timemoe}, ChronosBolt~\citep{ansari2024chronos}, TTM~\citep{ekambaram2024tiny}, Moirai~\citep{woo2024moirai}, Timer-XL~\citep{liu2024timer}, TimesFM-2.0~\citep{das2024decoder}, Chronos~\citep{ansari2024chronos}, and advanced full-shot deep learning models, including PatchTST~\citep{nie2023time}, DLinear~\citep{Zeng2022AreTE}, iTransformer~\citep{liu2023itransformer}, Pathformer~\citep{chen2024pathformer}, and TimesNet~\citep{wu2022timesnet}.

\subsection{Evaluation Datasets}
\label{app:eval_data}
We select datasets from diverse domains and with varying sampling frequencies as evaluation datasets.
The details are summarized in Table \ref{tab:eval_dataset}.
For the evaluation on the TSLib benchmark, the datasets were split into training, validation, and test sets. 
The split for the ETT and Weather datasets follows the configuration adopted by iTransformer~\citep{liu2023itransformer}. 
All other datasets use a 70\%/10\%/20\% ratio for the training, validation, and test sets, respectively.
\begin{table}[htbp]
    \centering
    \caption{Detailed descriptions of evaluation datasets.}
    \label{tab:eval_dataset}
    \resizebox{0.7\linewidth}{!}{
    \begin{tabular}{l *{5}{c}}
        \toprule
        Dataset & Domain & Frequency & \# Time Series & \# Target & \# Time points\\
        \midrule
        ETTh1 & Energy & H & 1 & 7 & 17,420
        \\
        ETTh2 & Energy & H & 1 & 7 & 17,420
        \\
        ETTm1 & Energy & 15T & 1 & 7 & 69,680
        \\
        ETTm2 & Energy & 15T & 1 & 7 & 69,680
        \\
        Weather & Nature & 10T & 1 & 21 & 52,696
        \\
        Saugeen (D) & Nature & D & 1 & 1 & 23,741
        \\
        Saugeen (W) & Nature & W & 1 & 1 & 3,391
        \\
        \bottomrule
    \end{tabular}}
\end{table}

\subsection{Evaluation Length Selection}
\label{app:Evaluation_Length_Selection}
In this section, we provide further details regarding the selection of historical sequence lengths for the various models evaluated in the TSLib benchmark, supplementing the discussion in Section \ref{sec:eval_datasets_and_metric}.

To accommodate diverse application scenarios, an increasing number of TSFMs~\citep{liu2025sundial, das2024decoder,liu2024timer,woo2024moirai} have devoted attention to predicting over long contexts. 
Consequently, we evaluate \method and other TSFMs under a long‑context setting. 
Specifically, we adopt a context length of 2048 time steps and examine four prediction horizons, which are \{96,192,336,720\}. 
For TSFMs incapable of processing this context length, we instead employ the context length at which each model achieves its best performance.

For the full-shot deep learning baselines, \textbf{a hyperparameter search was conducted independently for each dataset to ensure a fair evaluation}.
We compared the performance of each model using the context length from its original publication against a length of 2048, selecting the superior of the two.
To be specific, lengths of 96 and 2048 were compared for DLinear~\citep{Zeng2022AreTE}, iTransformer~\citep{liu2023itransformer}, TimesNet~\citep{wu2022timesnet}, and Pathformer, while lengths of 336, 512, and 2048 were compared for PatchTST~\citep{nie2023time}. 

The definitive context lengths adopted for each model are systematically tabulated in Table \ref{tab:eval_length}.

\begin{table}[htbp]
    \centering
    \caption{Context Lengths for Models on the TSLib Benchmark.}
    \label{tab:eval_length}
    
    \resizebox{\textwidth}{!}{ 
        \begin{tabular}{cccccccccccccc}
        \toprule
         Method & DLinear & iTrans. & TimesNet & PatchTST & Path. & Chronos & Moirai & TimesFM-2.0 & Timer-XL & $\text{TTM}_a$ & ChronosBolt & \method \\
        \midrule
        Context length & \{96, 2048\} & \{96, 2048\} & \{96, 2048\} & \{336, 512, 2048\} & \{96, 2048\} & 512 & 2048 & 2048 & 2048 & 1536 & 2048 & 2048 \\
        \bottomrule
        \end{tabular}
    }
\end{table}

\subsection{Details of DRoPE Analysis}
\label{sec:DRoPE_analysis}

In this section, we explain in more detail the setting of the DRoPE analysis experiment discussed in Section~\ref{exp:DRoPE_analysis}.
To more definitively ascertain whether these instance-specific modulations truly capture and leverage beneficial positional information derived from an instance's FFT features, we designed a series of \enquote{shuffle} experiments.
These experiments function as a form of causal intervention analysis. 
The underlying hypothesis is: if the instance-derived $\theta_{\text{inst}}$ modulations are crucial for DRoPE's performance, then disrupting this linkage by applying $\theta_{\text{inst}}$ modulations from one instance to another (shuffling) should lead to a noticeable degradation in forecasting accuracy. 

The details of each group in the experiment are as follows:
\begin{itemize}[leftmargin=5mm] 
    \item DRoPE: Our proposed method, where each time series instance utilizes its own FFT-derived features to independently modulate its RoPE $\theta_{\text{inst}}$ parameters at each layer.
    \item Intra-Dataset Shuffle: During inference, the learned layer-specific $\theta_{\text{inst}}$ modulation parameters ($\gamma^{(l)}, \beta^{(l)}$) are randomly permuted among instances within the same batch and originating from the same dataset. The modulated $\theta_{\text{inst}}$ are always shuffled in the same layer.
    \item Inter-Dataset Shuffle: For instances of a target dataset, layer-specific $\theta_{\text{inst}}$ modulation parameters ($\gamma^{(l)}, \beta^{(l)}$) are randomly sampled from a pre-computed collection derived from a different source dataset and applied to instances of the target dataset. This process also ensures layer-wise correspondence of the applied modulations.
    \item Fixed RoPE: To evaluate the architectural contribution of our module, this setting serves as a structural baseline trained from scratch. We replace the DRoPE module with standard RoPE, entirely removing the parameter-predicting MLP. All instances use the static, predefined RoPE $\theta$ across all layers during both training and inference.
\end{itemize}
By comparing DRoPE's performance against the inference-time causal interventions (shuffled configurations) and the fixed $\theta$ baseline, we can definitively attribute the performance gains to the effectiveness of the instance-adaptive $\theta_{\text{inst}}$ modulation process.

\section{Full Evaluation Results}
\label{sec:additional_results}
In this section, we present the detailed results in Section \ref{sec:zero_shot_eval}.
Table \ref{tab:full-zero-shot_performance} presents the full results of the zero-shot forecasting experiments conducted at each forecast horizon.
For Time-MoE, we report the results as presented in the original paper~\citep{shi2024timemoe}.

\begin{sidewaystable*}[bthp]
\caption{
Full results of zero-shot forecasting experiments.
AVG averages horizons within each dataset or all 27 dataset--horizon pairs overall.
Best results are highlighted in \textbf{bold}, and second best results are \underline{underlined}.
For fair comparison, the full-shot model is excluded from the ranking and is presented for reference only.
The dash (-) indicates that the dataset was utilized by the model during its pretraining phase. 
} 
\label{tab:full-zero-shot_performance}
\centering
\resizebox{\textheight}{!}{
\begin{tabular}{@{}l l l *{34}{c} c c} 
\toprule
\multirow{2}{*}{Method} 
& \multirow{2}{*}{Metric}
& \multicolumn{5}{c}{ETTh1} 
& \multicolumn{5}{c}{ETTh2}
& \multicolumn{5}{c}{ETTm1}
& \multicolumn{5}{c}{ETTm2}
& \multicolumn{5}{c}{Weather}
& \multicolumn{5}{c}{Saugeen(D)}
& \multicolumn{4}{c}{Saugeen(W)}
& \multirow{2}{*}{\textbf{AVG}}
& \multirow{2}{*}{Total $1^{\textbf{st}}$} \\
\cmidrule(lr){3-7} \cmidrule(lr){8-12} \cmidrule(lr){13-17} \cmidrule(lr){18-22} 
\cmidrule(lr){23-27} \cmidrule(lr){28-32} \cmidrule(lr){33-36}
& & 96 & 192 & 336 & 720 & AVG
& 96 & 192 & 336 & 720 & AVG
& 96 & 192 & 336 & 720 & AVG
& 96 & 192 & 336 & 720 & AVG
& 96 & 192 & 336 & 720 & AVG
& 96 & 192 & 336 & 720 & AVG
& 96 & 192 & 336 & AVG
& & \\
\midrule

\multirow{2}{*}{DLinear} 
 & MSE & 0.377 & 0.416 & 0.446 & 0.482 & 0.430 & 0.280 & 0.398 & 0.505 & 0.796 & 0.495 & 0.312 & 0.339 & 0.364 & 0.401 & 0.354 & 0.174 & 0.228 & 0.287 & 0.396 & 0.271 & 0.174 & 0.214 & 0.256 & 0.309 & 0.238 & 0.829 & 0.832 & 0.826 & 0.858 & 0.836 & 0.820 & 0.849 & 0.781 & 0.817 & 0.480 & \multirow{2}{*}{-}\\
 & MAE & 0.391 & 0.416 & 0.440 & 0.495 & 0.436 & 0.358 & 0.434 & 0.495 & 0.634 & 0.480 & 0.359 & 0.375 & 0.392 & 0.415 & 0.385 & 0.272 & 0.312 & 0.355 & 0.428 & 0.342 & 0.238 & 0.272 & 0.304 & 0.346 & 0.290 & 0.502 & 0.503 & 0.506 & 0.513 & 0.506 & 0.569 & 0.571 & 0.551 & 0.564 & 0.424 \\
\cmidrule(r){2-38}
\multirow{2}{*}{iTransformer} 
 & MSE & 0.383 & 0.441 & 0.478 & 0.470 & 0.443 & 0.282 & 0.367 & 0.402 & 0.406 & 0.364 & 0.327 & 0.363 & 0.390 & 0.424 & 0.376 & 0.336 & 0.369 & 0.398 & 0.428 & 0.383 & 0.171 & 0.218 & 0.278 & 0.357 & 0.256 & 0.778 & 0.795 & 0.800 & 0.839 & 0.803 & 0.988 & 1.041 & 1.007 & 1.012 & 0.501 & \multirow{2}{*}{-}\\
 & MAE & 0.396 & 0.429 & 0.449 & 0.469 & 0.436 & 0.344 & 0.392 & 0.422 & 0.433 & 0.398 & 0.355 & 0.381 & 0.411 & 0.442 & 0.397 & 0.355 & 0.381 & 0.411 & 0.442 & 0.397 & 0.210 & 0.256 & 0.299 & 0.351 & 0.279 & 0.457 & 0.463 & 0.482 & 0.490 & 0.473 & 0.626 & 0.640 & 0.642 & 0.636 & 0.423 \\
\cmidrule(r){2-38}
\multirow{2}{*}{TimesNet} 
 & MSE & 0.408 & 0.456 & 0.508 & 0.567 & 0.485 & 0.315 & 0.435 & 0.481 & 0.455 & 0.422 & 0.358 & 0.428 & 0.494 & 0.550 & 0.458 & 0.175 & 0.251 & 0.307 & 0.412 & 0.286 & 0.174 & 0.229 & 0.282 & 0.360 & 0.261 & 1.177 & 1.268 & 1.129 & 1.136 & 1.178 & 0.817 & 0.777 & 0.751 & 0.782 & 0.544 & \multirow{2}{*}{-}\\
 & MAE & 0.426 & 0.451 & 0.496 & 0.521 & 0.474 & 0.359 & 0.422 & 0.456 & 0.461 & 0.425 & 0.373 & 0.412 & 0.447 & 0.480 & 0.428 & 0.255 & 0.305 & 0.343 & 0.407 & 0.328 & 0.222 & 0.265 & 0.304 & 0.352 & 0.286 & 0.546 & 0.561 & 0.534 & 0.531 & 0.543 & 0.517 & 0.494 & 0.501 & 0.504 & 0.423 \\
\cmidrule(r){2-38}
\multirow{2}{*}{PatchTST} 
 & MSE & 0.382 & 0.416 & 0.443 & 0.466 & 0.427 & 0.280 & 0.351 & 0.388 & 0.424 & 0.361 & 0.278 & 0.327 & 0.363 & 0.414 & 0.346 & 0.162 & 0.221 & 0.270 & 0.369 & 0.256 & 0.149 & 0.195 & 0.254 & 0.346 & 0.236 & 0.827 & 0.83 & 0.819 & 0.848 & 0.831 & 0.849 & 0.847 & 0.781 & 0.826 & 0.456 & \multirow{2}{*}{-}\\
 & MAE & 0.403 & 0.426 & 0.444 & 0.474 & 0.437 & 0.349 & 0.393 & 0.420 & 0.445 & 0.402 & 0.330 & 0.361 & 0.388 & 0.423 & 0.376 & 0.249 & 0.291 & 0.323 & 0.386 & 0.312 & 0.205 & 0.248 & 0.293 & 0.354 & 0.275 & 0.474 & 0.482 & 0.488 & 0.491 & 0.484 & 0.545 & 0.526 & 0.500 & 0.524 & 0.397 \\
\cmidrule(r){2-38}
\multirow{2}{*}{Pathformer} 
 & MSE & 0.405 & 0.466 & 0.506 & 0.494 & 0.468 & 0.276 & 0.365 & 0.420 & 0.429 & 0.373 & 0.302 & 0.356 & 0.395 & 0.463 & 0.379 & 0.162 & 0.225 & 0.290 & 0.414 & 0.273 & 0.152 & 0.200 & 0.258 & 0.341 & 0.238 & 1.168 & 1.266 & 1.142 & 1.149 & 1.181 & 0.932 & 0.949 & 0.877 & 0.919 & 0.533 & \multirow{2}{*}{-}\\
 & MAE & 0.398 & 0.427 & 0.447 & 0.463 & 0.434 & 0.334 & 0.386 & 0.428 & 0.443 & 0.398 & 0.325 & 0.361 & 0.388 & 0.429 & 0.376 & 0.244 & 0.288 & 0.333 & 0.403 & 0.317 & 0.192 & 0.237 & 0.281 & 0.335 & 0.261 & 0.501 & 0.535 & 0.508 & 0.499 & 0.511 & 0.532 & 0.532 & 0.533 & 0.532 & 0.399 \\
\midrule

\multirow{2}{*}{$\text{Chronos}_{s}$} 
 & MSE & 0.516 & 0.599 & 0.613 & 0.603 & 0.583 & 0.304 & 0.407 & 0.427 & 0.450 & 0.397 & 0.504 & 0.601 & 0.667 & 0.728 & 0.625 & 0.207 & 0.277 & 0.343 & 0.468 & 0.324 & 0.209 & 0.267 & 0.328 & 0.397 & 0.300 & 1.108 & 1.142 & 1.105 & 1.143 & 1.125 & 1.183 & 1.227 & 1.165 & 1.192 & 0.629 & \multirow{2}{*}{0}\\
 & MAE & 0.416 & 0.464 & 0.487 & 0.512 & 0.470 & 0.347 & 0.401 & 0.428 & 0.457 & 0.408 & 0.398 & 0.457 & 0.498 & 0.538 & 0.473 & 0.277 & 0.323 & 0.364 & 0.437 & 0.350 & 0.213 & 0.265 & 0.307 & 0.356 & 0.285 & 0.477 & 0.503 & 0.506 & 0.512 & 0.500 & 0.520 & 0.524 & 0.526 & 0.523 & 0.426 \\
\cmidrule(r){2-38}
\multirow{2}{*}{$\text{Chronos}_{b}$} 
 & MSE & 0.526 & 0.585 & 0.593 & 0.608 & 0.578 & 0.322 & 0.411 & 0.440 & 0.443 & 0.404 & 0.422 & 0.523 & 0.600 & 0.686 & 0.558 & 0.206 & 0.272 & 0.329 & 0.436 & 0.311 & 0.184 & 0.241 & 0.296 & 0.375 & 0.274 & 1.172 & 1.224 & 1.202 & 1.249 & 1.212 & 1.166 & 1.222 & 1.157 & 1.182 & 0.626 & \multirow{2}{*}{0}\\
 & MAE & 0.413 & 0.447 & 0.461 & 0.491 & 0.453 & 0.343 & 0.392 & 0.422 & 0.440 & 0.399 & 0.374 & 0.434 & 0.477 & 0.520 & 0.451 & 0.270 & 0.316 & 0.354 & 0.420 & 0.340 & 0.206 & 0.256 & 0.298 & 0.353 & 0.278 & 0.473 & 0.505 & 0.502 & 0.505 & 0.496 & 0.515 & 0.517 & 0.508 & 0.513 & 0.415 \\
\cmidrule(r){2-38}
\multirow{2}{*}{$\text{Chronos}_{l}$} 
 & MSE & 0.493 & 0.601 & 0.636 & 0.653 & 0.596 & 0.328 & 0.434 & 0.480 & 0.522 & 0.442 & 0.452 & 0.536 & 0.613 & 0.671 & 0.568 & 0.203 & 0.268 & 0.316 & 0.424 & 0.303 & 0.195 & 0.254 & 0.311 & 0.395 & 0.289 & 1.178 & 1.229 & 1.194 & 1.232 & 1.208 & 1.213 & 1.269 & 1.241 & 1.241 & 0.642 & \multirow{2}{*}{0}\\
 & MAE & 0.400 & 0.449 & 0.475 & 0.503 & 0.457 & 0.348 & 0.407 & 0.448 & 0.482 & 0.421 & 0.379 & 0.430 & 0.470 & 0.512 & 0.448 & 0.263 & 0.309 & 0.342 & 0.408 & 0.331 & 0.211 & 0.264 & 0.305 & 0.360 & 0.285 & 0.481 & 0.508 & 0.499 & 0.499 & 0.497 & 0.523 & 0.534 & 0.544 & 0.534 & 0.420 \\
\cmidrule(r){2-38}
\multirow{2}{*}{$\text{Moirai}_{s}$} 
 & MSE & 0.395 & 0.430 & 0.445 & 0.430 & 0.425 & 0.274 & 0.341 & \underline{0.364} & \underline{0.378} & 0.339 & 0.362 & 0.385 & 0.410 & 0.450 & 0.402 & 0.192 & 0.258 & 0.307 & 0.391 & 0.287 & 0.160 & 0.207 & 0.258 & \textbf{0.327} & 0.238 & 1.082 & 1.112 & 1.085 & 1.105 & 1.096 & 1.272 & 1.284 & 1.177 & 1.244 & 0.551 & \multirow{2}{*}{4}\\
 & MAE & 0.404 & 0.427 & 0.436 & 0.445 & 0.428 & 0.324 & 0.369 & \textbf{0.386} & \textbf{0.405} & 0.371 & 0.360 & 0.379 & 0.399 & 0.422 & 0.390 & 0.263 & 0.308 & 0.340 & 0.397 & 0.327 & 0.194 & 0.238 & 0.275 & \textbf{0.324} & 0.258 & 0.479 & 0.493 & 0.491 & 0.493 & 0.489 & 0.560 & 0.562 & 0.535 & 0.552 & 0.397 \\
\cmidrule(r){2-38}
\multirow{2}{*}{$\text{Moirai}_{b}$} 
 & MSE & 0.429 & 0.494 & 0.531 & 0.544 & 0.500 & 0.275 & 0.347 & 0.379 & 0.388 & 0.347 & 0.456 & 0.459 & 0.482 & 0.518 & 0.479 & 0.194 & 0.251 & 0.292 & \textbf{0.354} & 0.273 & 0.170 & 0.218 & 0.270 & 0.344 & 0.251 & 1.129 & 1.175 & 1.142 & 1.177 & 1.156 & 1.208 & 1.243 & 1.171 & 1.207 & 0.579 & \multirow{2}{*}{2}\\
 & MAE & 0.412 & 0.447 & 0.467 & 0.501 & 0.457 & 0.319 & 0.365 & \underline{0.388} & 0.408 & 0.370 & 0.385 & 0.400 & 0.415 & 0.435 & 0.409 & 0.266 & 0.306 & 0.333 & \textbf{0.375} & 0.320 & 0.203 & 0.249 & 0.288 & 0.339 & 0.270 & 0.472 & 0.495 & 0.485 & 0.485 & 0.484 & 0.557 & 0.550 & 0.538 & 0.548 & 0.403 \\
\cmidrule(r){2-38}
\multirow{2}{*}{$\text{Moirai}_{l}$} 
 & MSE & 0.464 & 0.523 & 0.546 & 0.549 & 0.521 & 0.298 & 0.403 & 0.464 & 0.530 & 0.424 & 0.522 & 0.541 & 0.563 & 0.611 & 0.559 & 0.195 & 0.257 & 0.308 & 0.395 & 0.289 & 0.170 & 0.222 & 0.277 & 0.361 & 0.258 & 1.102 & 1.140 & 1.102 & 1.122 & 1.117 & 1.213 & 1.252 & 1.179 & 1.215 & 0.604 & \multirow{2}{*}{0}\\
 & MAE & 0.433 & 0.470 & 0.486 & 0.511 & 0.475 & 0.346 & 0.412 & 0.448 & 0.498 & 0.426 & 0.401 & 0.423 & 0.442 & 0.465 & 0.433 & 0.267 & 0.311 & 0.342 & 0.395 & 0.329 & 0.203 & 0.253 & 0.293 & 0.347 & 0.274 & 0.464 & 0.486 & 0.484 & 0.486 & 0.480 & 0.541 & 0.543 & 0.534 & 0.539 & 0.418 \\
\cmidrule(r){2-38}
\multirow{2}{*}{TimesFM-2.0} 
 & MSE & 0.424 & 0.444 & 0.443 & 0.438 & 0.437 & \underline{0.256} & \underline{0.328} & 0.366 & 0.391 & \underline{0.335} & 0.362 & 0.410 & 0.429 & 0.470 & 0.418 & 0.175 & 0.239 & 0.294 & 0.369 & 0.269 & - & - & - & - & - & 1.099 & 1.106 & 1.058 & 1.088 & 1.088 & 1.277 & 1.448 & 3.971 & 2.232 & - & \multirow{2}{*}{0}\\
 & MAE & 0.397 & 0.417 & 0.424 & 0.442 & 0.420 & 0.308 & \underline{0.361} & 0.389 & 0.413 & \underline{0.368} & 0.340 & 0.373 & 0.395 & 0.424 & 0.383 & 0.245 & 0.294 & 0.332 & 0.385 & 0.314 & - & - & - & - & - & 0.449 & 0.483 & 0.492 & 0.522 & 0.487 & 0.649 & 0.724 & 1.362 & 0.912 & - \\
\cmidrule(r){2-38}
\multirow{2}{*}{Timer-XL} 
 & MSE & 0.394 & 0.436 & 0.469 & 0.436 & 0.434 & 0.273 & 0.342 & 0.377 & 0.382 & 0.344 & 0.325 & 0.374 & 0.414 & 0.510 & 0.406 & 0.187 & 0.240 & 0.285 & 0.374 & 0.272 & 0.167 & 0.213 & 0.266 & 0.340 & 0.247 & 1.043 & 1.126 & 1.127 & 1.127 & 1.106 & 1.112 & 1.128 & 1.043 & 1.094 & 0.537 & \multirow{2}{*}{0}\\
 & MAE & 0.395 & 0.419 & 0.437 & 0.448 & 0.425 & 0.336 & 0.380 & 0.402 & 0.416 & 0.384 & 0.354 & 0.385 & 0.409 & 0.455 & 0.401 & 0.272 & 0.310 & 0.338 & 0.392 & 0.328 & 0.221 & 0.264 & 0.304 & 0.354 & 0.286 & 0.574 & 0.612 & 0.615 & 0.606 & 0.602 & 0.626 & 0.620 & 0.594 & 0.613 & 0.427 \\
\cmidrule(r){2-38}
\multirow{2}{*}{$\text{Time-MoE}_{b}$} 
 & MSE & \underline{0.357} & \textbf{0.384} & \underline{0.411} & 0.449 & 0.400 & 0.305 & 0.351 & 0.391 & 0.419 & 0.367 & 0.338 & 0.353 & 0.381 & 0.504 & 0.394 & 0.201 & 0.258 & 0.324 & 0.488 & 0.318 & 0.160 & 0.210 & 0.274 & 0.418 & 0.266 & - & - & - & - & - & - & - & - & - & - & \multirow{2}{*}{3}\\
 & MAE & \textbf{0.381} & \textbf{0.404} & 0.434 & 0.477 & 0.424 & 0.359 & 0.386 & 0.418 & 0.454 & 0.404 & 0.368 & 0.388 & 0.413 & 0.493 & 0.416 & 0.291 & 0.334 & 0.373 & 0.464 & 0.366 & 0.214 & 0.260 & 0.309 & 0.405 & 0.297 & - & - & - & - & - & - & - & - & - & - \\
\cmidrule(r){2-38}
\multirow{2}{*}{$\text{Time-MoE}_{l}$} 
 & MSE & \textbf{0.350} & \underline{0.388} & \underline{0.411} & 0.427 & \textbf{0.394} & 0.302 & 0.364 & 0.417 & 0.537 & 0.405 & 0.309 & \underline{0.346} & \underline{0.373} & 0.475 & 0.376 & 0.197 & 0.250 & 0.337 & 0.480 & 0.316 & 0.159 & 0.215 & 0.291 & 0.415 & 0.270 & - & - & - & - & - & - & - & - & - & - & \multirow{2}{*}{2}\\
 & MAE & \underline{0.382} & 0.412 & 0.430 & 0.455 & 0.420 & 0.354 & 0.385 & 0.425 & 0.496 & 0.415 & 0.357 & 0.381 & 0.408 & 0.477 & 0.406 & 0.286 & 0.322 & 0.375 & 0.461 & 0.361 & 0.213 & 0.266 & 0.322 & 0.400 & 0.300 & - & - & - & - & - & - & - & - & - & - \\
\cmidrule(r){2-38}
\multirow{2}{*}{$\text{TTM}_{a}$} 
 & MSE & 0.373 & 0.396 & \textbf{0.406} & \textbf{0.411} & \underline{0.397} & 0.262 & 
 \textbf{0.326} & \textbf{0.354} & \textbf{0.376} & \textbf{0.330} & 0.329 & 0.364 & 0.381 & 0.413 & 0.372 & 0.167 & 0.227 & \underline{0.277} & \underline{0.360} & \underline{0.258} & \underline{0.146} & \underline{0.193} & \underline{0.249} & \underline{0.328} & \textbf{0.229} & 0.929 & \textbf{0.926} & \textbf{0.904} & \textbf{0.939} & \textbf{0.925} & 1.114 & 1.134 & 1.051 & 1.100 & 0.494 & \multirow{2}{*}{11}\\
 & MAE & 0.397 & 0.415 & 0.424 & 0.441 & 0.419 & 0.334 & 0.378 & 0.400 & 0.421 & 0.383 & 0.350 & 0.374 & 0.390 & 0.410 & 0.381 & 0.254 & 0.296 & 0.329 & 0.382 & 0.315 & 0.197 & 0.242 & 0.284 & 0.336 & 0.265 & 0.511 & 0.518 & 0.523 & 0.541 & 0.523 & 0.628 & 0.628 & 0.608 & 0.621 & 0.408 \\
\cmidrule(r){2-38}
\multirow{2}{*}{$\text{ChronosBolt}_{s}$} 
 & MSE & 0.405 & 0.475 & 0.514 & 0.516 & 0.478 & \textbf{0.255} & 0.339 & 0.388 & 0.400 & 0.346 & 0.309 & 0.375 & 0.420 & 0.512 & 0.404 & 0.166 & 0.231 & 0.289 & 0.400 & 0.272 & 0.152 & 0.205 & 0.264 & 0.357 & 0.245 & \textbf{0.863} & 0.956 & 1.001 & 1.103 & 0.981 & \underline{0.762} & \underline{0.802} & \underline{0.749} & \underline{0.771} & 0.489 & \multirow{2}{*}{2} \\
 & MAE & 0.388 & 0.426 & 0.453 & 0.485 & 0.438 & \underline{0.307} & 0.362 & 0.398 & 0.418 & 0.371 & \underline{0.317} & 0.357 & 0.389 & 0.441 & 0.376 & \underline{0.237} & 0.284 & 0.324 & 0.393 & 0.310 & 0.186 & 0.237 & 0.279 & 0.338 & 0.260 & \underline{0.410} & 0.458 & 0.492 & 0.551 & 0.478 & \underline{0.433} & 0.446 & 0.473 & 0.451 & 0.381 \\
\cmidrule(r){2-38}
\multirow{2}{*}{$\text{ChronosBolt}_{b}$} 
 & MSE & 0.420 & 0.486 & 0.517 & 0.493 & 0.479 & \textbf{0.255} & 0.333 & 0.381 & 0.393 & 0.341 & \underline{0.303} & 0.370 & 0.410 & 0.497 & 0.395 & \underline{0.164} & 0.233 & 0.299 & 0.414 & 0.278 & 0.150 & 0.197 & 0.255 & 0.344 & 0.237 & \underline{0.905} & 0.959 & 0.957 & 1.049 & 0.968 & \textbf{0.752} & \textbf{0.791} & \textbf{0.725} & \textbf{0.756} & 0.483 & \multirow{2}{*}{\underline{14}}\\
 & MAE & 0.388 & 0.424 & 0.445 & 0.457 & 0.429 & \textbf{0.305} & \textbf{0.355} & 0.389 & 0.408 & \textbf{0.364} & \textbf{0.311} & \textbf{0.352} & \underline{0.382} & 0.428 & \underline{0.368} & \textbf{0.232} & \textbf{0.280} & 0.323 & 0.391 & \underline{0.307} & \underline{0.183} & 0.230 & \underline{0.272} & 0.329 & \underline{0.254} & 0.417 & 0.439 & 0.441 & 0.467 & 0.441 & \textbf{0.423} &\textbf{0.437} & 0.462 & 0.441 & 0.369 \\
\cmidrule(r){2-38}
\multirow{2}{*}{$\method_{m}$} 
 & MSE & 0.386 & 0.436 & 0.460 & 0.426 & 0.427 & 0.257 & 0.332 & 0.369 & 0.382 & \underline{0.335} & 0.304 & 0.358 & 0.383 & \underline{0.408} & 0.363 & 0.168 & 0.228 & 0.279 & 0.366 & 0.260 & \textbf{0.144} & \underline{0.193} & 0.254 & 0.353 & 0.236 & 0.976 & 0.997 & 0.982 & 1.015 & 0.993 & 0.789 & 0.840 & 0.788 & 0.806 & 0.477 & \multirow{2}{*}{4}\\
 & MAE & 0.384 & 0.411 & \underline{0.422} & \underline{0.425} & \underline{0.411} & 0.314 & 0.364 & 0.391 & \underline{0.407} & 0.369 & 0.329 & 0.362 & 0.384 & 0.407 & 0.371 & 0.246 & 0.290 & 0.323 & 0.378 & 0.309 & \textbf{0.181} & \underline{0.229} & 0.273 & 0.331 & \underline{0.254} & 0.424 & 0.439 & 0.443 & 0.450 & 0.439 & 0.437 & \underline{0.444} & \textbf{0.426} & \textbf{0.436} & 0.367 \\
\cmidrule(r){2-38}
\multirow{2}{*}{$\method_{s}$} 
 & MSE & 0.385 & 0.429 & 0.446 & \underline{0.422} & 0.421 & 0.269 & 0.340 & 0.380 & 0.394 & 0.346 & 0.305 & 0.356 & 0.381 & \textbf{0.404} & \underline{0.362} & \textbf{0.163} & \underline{0.226} & 0.278 & 0.382 & 0.262 & 0.151 & 0.197 & 0.253 & 0.340 & 0.235 & 0.920 & \underline{0.943} & \underline{0.942} & \underline{1.002} & \underline{0.952} & 0.808 & 0.851 & 0.815 & 0.825 & \underline{0.473} & \multirow{2}{*}{6}\\
 & MAE & 0.383 & \underline{0.408} & \textbf{0.418} & \textbf{0.424} & \textbf{0.408} & 0.320 & 0.368 & 0.397 & 0.416 & 0.375 & 0.326 & 0.360 & \underline{0.382} & \textbf{0.404} & \underline{0.368} & 0.238 & 0.284 & \underline{0.320} & 0.384 & \underline{0.307} & 0.186 & 0.232 & 0.274 & 0.329 & 0.255 & \underline{0.410} & \underline{0.426} & \underline{0.429} & \underline{0.442} & \underline{0.427} & 0.447 & 0.446 & \underline{0.429} & 0.441 & \underline{0.366}\\
\cmidrule(r){2-38}
\multirow{2}{*}{$\method_{b}$} 
 & MSE & 0.396 & 0.446 & 0.472 & 0.434 & 0.437 & 0.270 & 0.348 & 0.367 & \textbf{0.376} & 0.340 & \textbf{0.291} & \textbf{0.337}  & \textbf{0.368} & \underline{0.408} & \textbf{0.351} & \textbf{0.163} & \textbf{0.224} & \textbf{0.274} & 0.361 & \textbf{0.255} & 0.147 & \textbf{0.191} & \textbf{0.248} & 0.333 & \underline{0.230} & 0.936 & 0.972 & 0.963 & 1.014 & 0.971 & 0.786 & 0.817 & 0.766 & 0.790 & \textbf{0.471} & \multirow{2}{*}{\textbf{26}} \\    
 & MAE & 0.385 & 0.414 & 0.429 & 0.432 & 0.415 & 0.325 & 0.378 & 0.395 & 0.410 & 0.377 & 0.321 & \underline{0.355} & \textbf{0.378} & \underline{0.405} & \textbf{0.365} & 0.239 & \underline{0.283} & \textbf{0.317} & \underline{0.376} & \textbf{0.304} & \textbf{0.181} & \textbf{0.227} & \textbf{0.270} & \underline{0.326} & \textbf{0.251} & \textbf{0.406} & \textbf{0.424} & \textbf{0.428} & \textbf{0.437} & \textbf{0.424} & 0.445 & 0.446 & \underline{0.429} & \underline{0.440} & \textbf{0.365} \\
\bottomrule
\end{tabular}}
\end{sidewaystable*}

\section{Statistical Analysis and Training Variability}
\label{app:statistical_analysis}
We assess the robustness of the GIFT-Eval results through paired statistical comparisons, an analysis of gain and loss magnitudes, and repeated training runs.

\textbf{Dataset-Level Paired Tests.}
The 97 GIFT-Eval tasks originate from 28 datasets evaluated at different frequencies and forecast horizons.
To avoid treating correlated variants as independent observations, we group tasks by source dataset.
For each comparison, we compute the log ratio of the MASE of \method-Base to that of the compared model for each task and average these log ratios within each source dataset.
A negative dataset-level log ratio indicates a win for \method-Base.
We apply two-sided exact sign tests and two-sided exact Wilcoxon signed-rank tests to the same 28 paired dataset-level log ratios and report both tests side by side.
Holm correction is applied across the four comparisons separately within each test family.
The sign test uses win/loss directions, whereas the Wilcoxon test also incorporates the ranked magnitudes of the paired differences and assumes a symmetric distribution of those differences.
We additionally report Hodges--Lehmann estimates of the typical dataset-level MASE reduction and their 95\% confidence intervals.

\begin{table}[htbp]
\centering
\caption{Dataset-level statistical comparisons on GIFT-Eval using two-sided exact tests.
Wins and losses are counted across 28 source datasets.
Holm correction covers the four comparisons within each test family separately.
HL denotes the Hodges--Lehmann reduction estimate, reported with its 95\% confidence interval.}
\label{tab:statistical_tests}
\resizebox{\textwidth}{!}{%
\begin{tabular}{lcccccc}
\toprule
\textbf{Comparison} & \textbf{Wins/Losses} & \makecell{\textbf{Sign test}\\\textbf{$p$}} & \makecell{\textbf{Sign test}\\\textbf{Holm $p$}} & \makecell{\textbf{HL reduction}\\\textbf{[95\% CI]}} & \makecell{\textbf{Wilcoxon}\\\textbf{$p$}} & \makecell{\textbf{Wilcoxon}\\\textbf{Holm $p$}} \\
\midrule
\method-Base vs. Sundial & 20/8 & 0.0357 & 0.0714 & 3.45\% [0.14\%, 7.98\%] & 0.0402 & 0.0402 \\
Full vs. fixed patch & 19/9 & 0.0872 & 0.0872 & 1.06\% [0.10\%, 2.51\%] & 0.0179 & 0.0358 \\
Full vs. no null experts & 23/5 & 0.00091 & 0.00365 & 0.91\% [0.32\%, 2.17\%] & 0.0011 & 0.0032 \\
Full vs. vanilla RoPE & 23/5 & 0.00091 & 0.00365 & 2.62\% [1.42\%, 4.80\%] & 0.00022 & 0.00087 \\
\bottomrule
\end{tabular}%
}
\end{table}

At the 0.05 significance level after Holm correction, both tests are significant for the no-null-expert and vanilla-RoPE ablations.
For \method-Base versus Sundial and the full model versus fixed patching, the sign tests are non-significant, while the Wilcoxon tests are significant.
These outcomes provide complementary evidence: improvements over Sundial and fixed patching are significant when ranked effect magnitudes are included, but not when considering win/loss directions alone.

\textbf{Magnitudes of Gains and Losses.}
To characterize practical performance differences, Table~\ref{tab:gain_loss_magnitudes} reports both improvements and degradations relative to Sundial.
At the task level, \method wins on 64 of 97 tasks, with an 8.1\% geometric-average MASE reduction among wins; the remaining 33 tasks show a 12.3\% geometric-average worsening.
After grouping by source dataset, \method wins on 20 of 28 datasets, with a 9.36\% geometric-average reduction among wins, and loses on eight, with an 8.02\% worsening.
These percentages summarize the winning and losing groups separately and are distinct from the overall benchmark improvement.
Across all tasks, \method-Base achieves a lower normalized MASE than Sundial, 0.738 versus 0.750, corresponding to a 1.6\% reduction.
Thus, the overall advantage coexists with meaningful variation across tasks and datasets.

\begin{table}[htbp]
\centering
\caption{Magnitudes of gains and losses for \method-Base relative to Sundial.
Reduction and worsening are geometric-average changes within the winning and losing groups, respectively.}
\label{tab:gain_loss_magnitudes}
\begin{tabular}{lcccc}
\toprule
\textbf{Aggregation} & \textbf{Wins} & \textbf{Losses} & \makecell{\textbf{Reduction}\\\textbf{among wins}} & \makecell{\textbf{Worsening}\\\textbf{among losses}} \\
\midrule
Task & 64 & 33 & 8.1\% & 12.3\% \\
Source dataset & 20 & 8 & 9.36\% & 8.02\% \\
\bottomrule
\end{tabular}
\end{table}

\textbf{Repeated Training Runs.}
We train \method-Small independently three times using different random seeds.
Its GIFT-Eval normalized MASE is $0.747\pm0.001$, reported as the mean $\pm$ standard deviation across the three runs.
The small standard deviation indicates consistent performance across repeated training runs.
This measures training-run variability, complementing the paired tests of performance differences across datasets.
Inference is deterministic because \method directly predicts quantiles without stochastic sampling.

\section{Few-Shot Forecasting}
\label{app:few_shot}
To evaluate adaptation with limited target-domain data, we fine-tune \method-Base on ETTh1, ETTh2, ETTm1, ETTm2, and Weather.
For each dataset, we randomly sample 1\% or 10\% of the training windows from its training split and update all model parameters.
We report MSE and MAE averaged over prediction horizons $\{96,192,336,720\}$ and compare against the zero-shot model and full-shot PatchTST~\citep{nie2023time} and DLinear~\citep{Zeng2022AreTE}.

\begin{table}[htbp]
\centering
\caption{Few-shot forecasting results for \method-Base.
Each entry reports MSE / MAE averaged over prediction horizons $\{96,192,336,720\}$.
The percentages indicate the fraction of randomly sampled training windows used for full-parameter fine-tuning.
PatchTST and DLinear use the full training split.}
\label{tab:few_shot}
\resizebox{0.8\textwidth}{!}{%
\begin{tabular}{lccccc}
\toprule
& \multicolumn{3}{c}{\textbf{\method-Base}} & \multicolumn{2}{c}{\textbf{Full-shot baselines}} \\
\cmidrule(lr){2-4}\cmidrule(lr){5-6}
\textbf{Dataset} & \textbf{Zero-shot} & \textbf{1\% few-shot} & \textbf{10\% few-shot} & \textbf{PatchTST} & \textbf{DLinear} \\
\midrule
ETTh1 & 0.437 / 0.415 & 0.405 / 0.413 & \textbf{0.398 / 0.412} & 0.427 / 0.437 & 0.430 / 0.436 \\
ETTh2 & 0.340 / 0.377 & 0.328 / 0.367 & \textbf{0.319 / 0.363} & 0.361 / 0.402 & 0.495 / 0.480 \\
ETTm1 & 0.351 / 0.365 & 0.321 / 0.352 & \textbf{0.312 / 0.351} & 0.346 / 0.376 & 0.354 / 0.385 \\
ETTm2 & 0.255 / 0.304 & 0.233 / 0.293 & \textbf{0.230 / 0.290} & 0.256 / 0.312 & 0.271 / 0.342 \\
Weather & 0.230 / 0.251 & 0.210 / 0.245 & \textbf{0.208 / 0.239} & 0.236 / 0.275 & 0.238 / 0.290 \\
\bottomrule
\end{tabular}%
}
\end{table}

As shown in Table~\ref{tab:few_shot}, fine-tuning with only 1\% of training windows improves both MSE and MAE over zero-shot forecasting on all five datasets and surpasses both full-shot baselines.
Increasing the fraction to 10\% further improves both metrics on every dataset.
These results demonstrate that \method can effectively adapt its pre-trained representations using limited target-domain supervision.

\section{Calculation of information density}
\label{sec:cal_information_density}
In this section, we present the formula for the information density illustrated in Figure \ref{fig:intro}(b).
To effectively characterize and compare different time series datasets, we introduce a method to quantify their information density and the variation of this density over time.
For this, similar to TimeMixer~\citep{wang2024timemixer}, we utilize spectral entropy, a metric that measures the complexity and compressibility of a signal. 
In this context, a signal with low spectral entropy, such as one with a few dominant periodic components, is considered to have low information density due to its redundant and predictable nature.
Conversely, a signal with high spectral entropy, resembling white noise, has its power spread broadly across the frequency spectrum, indicating a high degree of randomness and thus a higher information density.

Given the observations $\mathbf{x}_{1:T} = (x_1, \ldots, x_T) \in \mathbb{R}^T$ of a specific dataset, we analyze the time series using a sliding window approach.
The series is segmented into $N$ windows, $\mathbf{w}_i$, each of size $M$ with a step size of $S$.
In our analysis, we use a non-overlapping configuration where both the window size and the step size are set to 128 (i.e., $M=128$, $S=128$).
The total number of windows is $N = \lfloor (T - M) / S \rfloor + 1$. 
To mitigate spectral leakage from windowing, we apply a Hann window to each segment $\mathbf{w}_i$:
\begin{equation}
    h[n] = 0.5 - 0.5\cos\left(\frac{2\pi n}{M-1}\right), \quad 0 \le n \le M-1.
\end{equation}

For each resulting windowed segment $\mathbf{w}'_i = \mathbf{w}_i \cdot h$, we compute its normalized power spectral density, $p_i[k]$, which describes how the signal's power is distributed over different frequencies. 
The spectral entropy for the $i$-th window, $H_{SE}(\mathbf{w}_i)$, is then calculated using the Shannon entropy formula:
\begin{equation}
    H_{SE}(\mathbf{w}_i) = -\sum_{k=0}^{M-1} p_i[k] \log_2(p_i[k]).
\end{equation}
This process yields a sequence of entropy values, $(H_{SE}(\mathbf{w}_1), \ldots, H_{SE}(\mathbf{w}_N))$, where each value represents the localized information density of its corresponding time segment.

To obtain a holistic view of the dataset's characteristics, we compute two key statistics from this entropy sequence.

First, the mean spectral entropy ($\mu_{SE}$) serves as a measure of the average information density of the entire dataset.
A higher $\mu_{SE}$ suggests that the dataset, on average, contains more complex and less predictable patterns. 
This allows for a direct comparison of the overall information content between different datasets.
\begin{equation}
    \mu_{SE} = \frac{1}{N} \sum_{i=1}^{N} H_{SE}(\mathbf{w}_i).
\end{equation}

Second, the standard deviation of spectral entropy ($\sigma_{SE}$) quantifies the variability of information density within the dataset.
A small $\sigma_{SE}$ indicates that the dataset is stationary in its complexity, with a consistent level of information density throughout. 
A large $\sigma_{SE}$, however, reveals a non-stationary character, signifying substantial fluctuations in the signal's complexity and information content over time.
\begin{equation}
    \sigma_{SE} = \sqrt{\frac{1}{N} \sum_{i=1}^{N} (H_{SE}(\mathbf{w}_i) - \mu_{SE})^2}.
\end{equation}
Together, $\mu_{SE}$ and $\sigma_{SE}$ provide a concise yet powerful summary of a dataset's informational characteristics, enabling a quantitative assessment of its average complexity and internal dynamics.


\section{Showcases}
\subsection{Showcases of Synthetic Data}
\label{app:synthetic_cases}
\begin{figure}[htbp]
    \centering
    \includegraphics[width=0.85\textwidth]{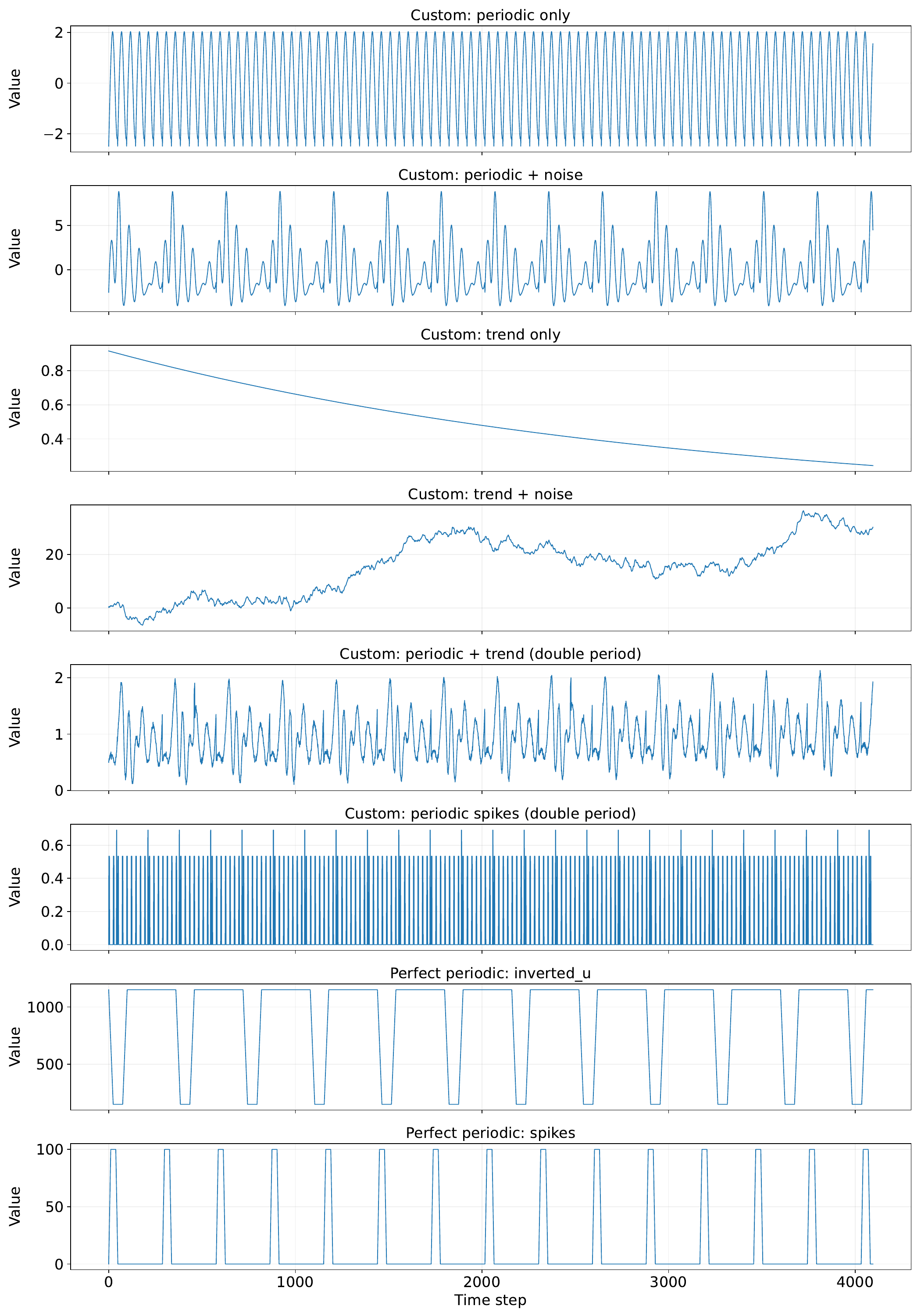}
    \caption{Several distinct case types are generated by the synthetic dataset algorithm. Specifically, the Custom designation refers to a composite signal type, which is systematically constructed by combining seasonal, trend, and noise components. Concurrently, the Perfect periodic designation denotes synthesized, idealized industrial signals.}
    \label{fig:syn_example}
\end{figure}

Figure~\ref{fig:syn_example}  showcases representative examples generated by the algorithm using the synthetic dataset.
As detailed in Appendix \ref{app:data}, the generated synthetic dataset is classified into two categories. 
The first is a composite type, formed from a combination of seasonal, trend, and noise components, which is designated Custom in the figures.
The second consists of idealized industrial signals, designated Perfect periodic.

\subsection{Showcases of \method}
\begin{figure}[htbp]
    \centering
    \captionsetup[subfigure]{labelformat=parens}

    \newlength{\myfigheight}
    \setlength{\myfigheight}{0.13\textheight}
    \begin{subfigure}[b]{0.49\textwidth}
        \centering
        \includegraphics[width=\linewidth,height=\myfigheight,keepaspectratio]{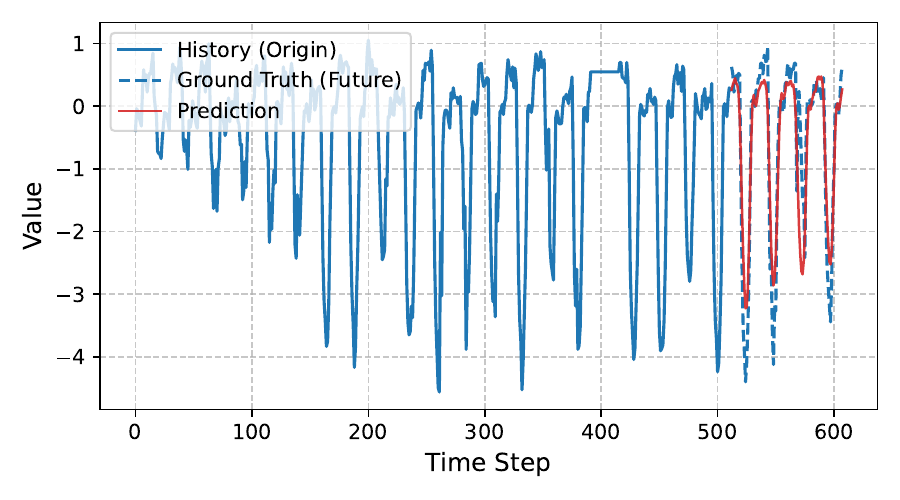}
        \caption{ETTh1-1}
        \label{fig:sub_a}
    \end{subfigure}
    \hfill
    \begin{subfigure}[b]{0.49\textwidth}
        \centering
        \includegraphics[width=\linewidth,height=\myfigheight,keepaspectratio]{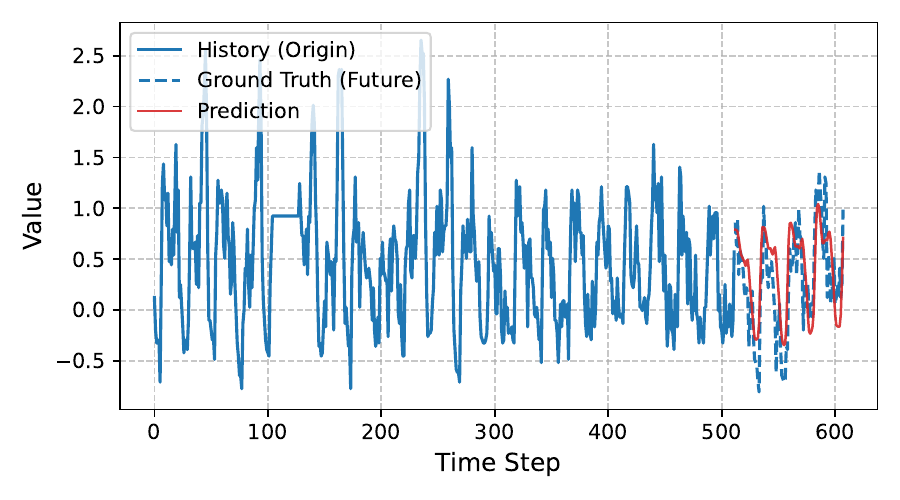}
        \caption{ETTh1-2}
        \label{fig:sub_b}
    \end{subfigure}

    \vspace{0.1cm}

    \begin{subfigure}[b]{0.49\textwidth}
        \centering
        \includegraphics[width=\linewidth,height=\myfigheight,keepaspectratio]{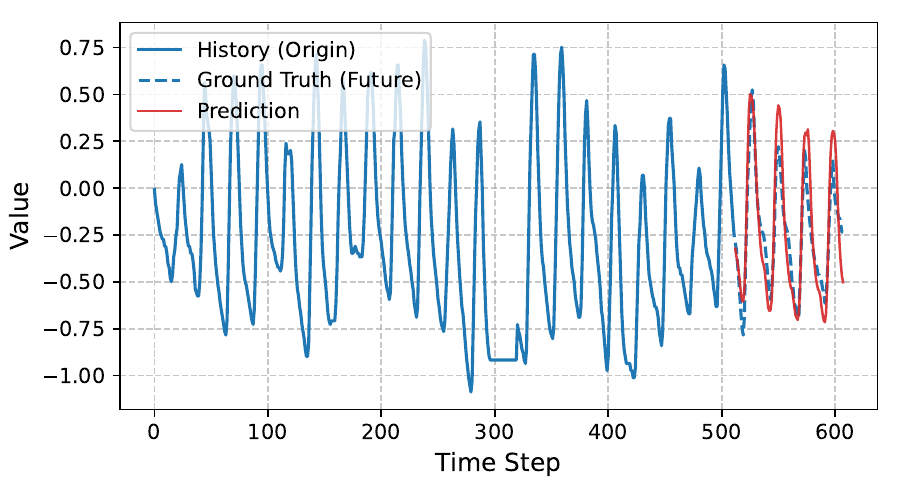}
        \caption{ETTh2-1}
        \label{fig:sub_c}
    \end{subfigure}
    \hfill
    \begin{subfigure}[b]{0.49\textwidth}
        \centering
        \includegraphics[width=\linewidth,height=\myfigheight,keepaspectratio]{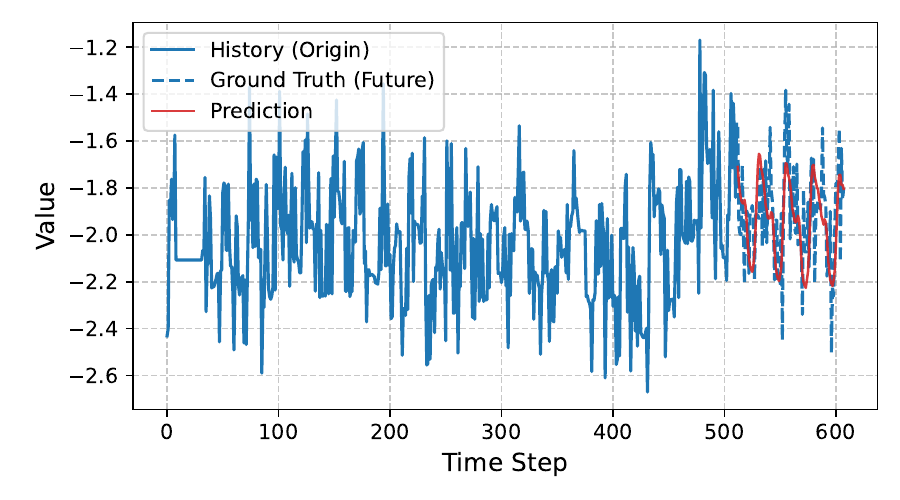}
        \caption{ETTh2-2}
        \label{fig:sub_d}
    \end{subfigure}

    \vspace{0.1cm}

    \begin{subfigure}[b]{0.49\textwidth}
        \centering
        \includegraphics[width=\linewidth,height=\myfigheight,keepaspectratio]{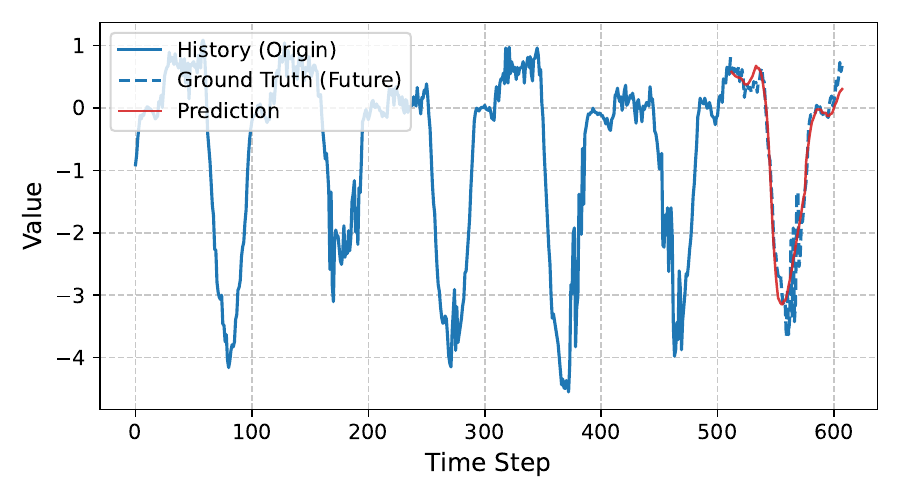}
        \caption{ETTm1-1}
        \label{fig:sub_e}
    \end{subfigure}
    \hfill
    \begin{subfigure}[b]{0.49\textwidth}
        \centering
        \includegraphics[width=\linewidth,height=\myfigheight,keepaspectratio]{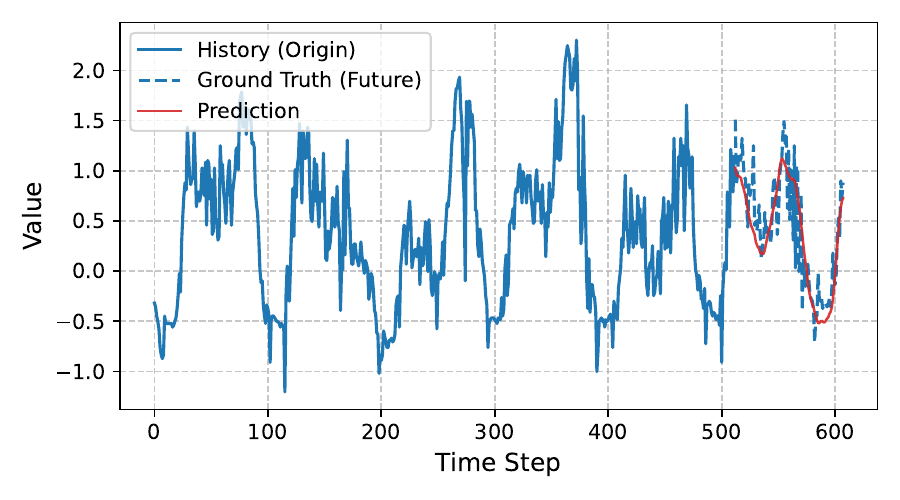}
        \caption{ETTm1-2}
        \label{fig:sub_f}
    \end{subfigure}

    \vspace{0.1cm}

    \begin{subfigure}[b]{0.49\textwidth}
        \centering
        \includegraphics[width=\linewidth,height=\myfigheight,keepaspectratio]{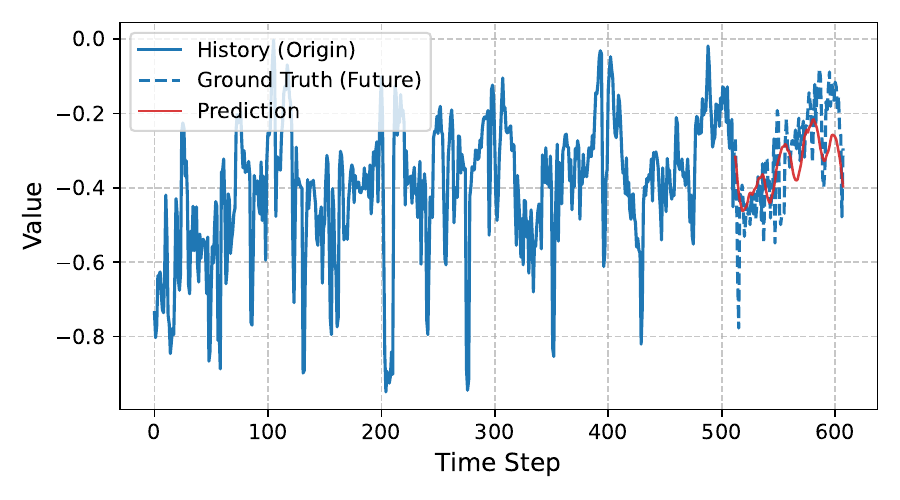}
        \caption{ETTm2-1}
        \label{fig:sub_g}
    \end{subfigure}
    \hfill
    \begin{subfigure}[b]{0.49\textwidth}
        \centering
        \includegraphics[width=\linewidth,height=\myfigheight,keepaspectratio]{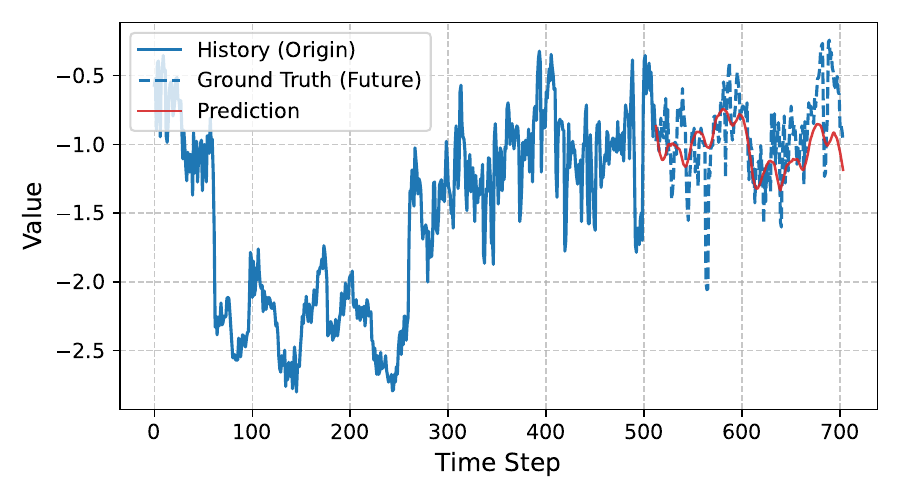}
        \caption{ETTm2-2}
        \label{fig:sub_h}
    \end{subfigure}

    \vspace{0.1cm}

    \begin{subfigure}[b]{0.49\textwidth}
        \centering
        \includegraphics[width=\linewidth,height=\myfigheight,keepaspectratio]{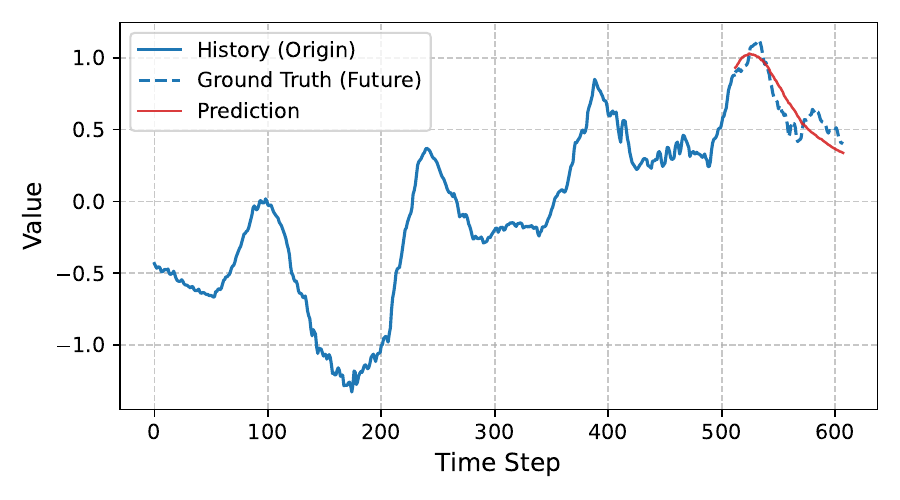}
        \caption{Weather-1}
        \label{fig:sub_i}
    \end{subfigure}
    \hfill
    \begin{subfigure}[b]{0.49\textwidth}
        \centering
        \includegraphics[width=\linewidth,height=\myfigheight,keepaspectratio]{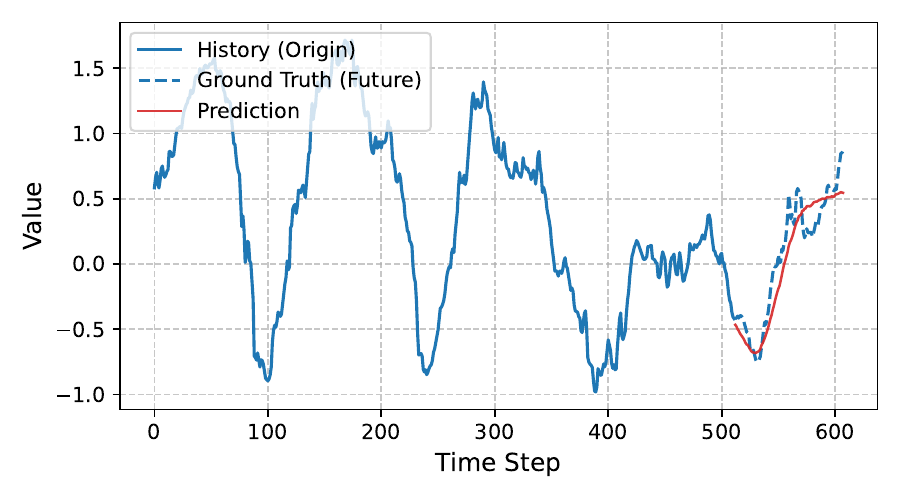}
        \caption{Weather-2}
        \label{fig:sub_j}
    \end{subfigure}

    \vspace{0.1cm}

    \begin{subfigure}[b]{0.49\textwidth}
        \centering
        \includegraphics[width=\linewidth,height=\myfigheight,keepaspectratio]{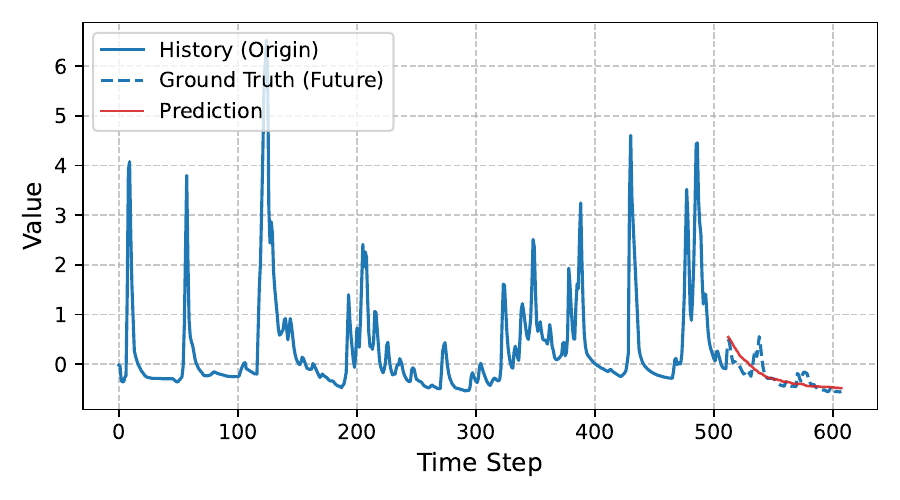}
        \caption{Saugeen-1}
        \label{fig:sub_k}
    \end{subfigure}
    \hfill
    \begin{subfigure}[b]{0.49\textwidth}
        \centering
        \includegraphics[width=\linewidth,height=\myfigheight,keepaspectratio]{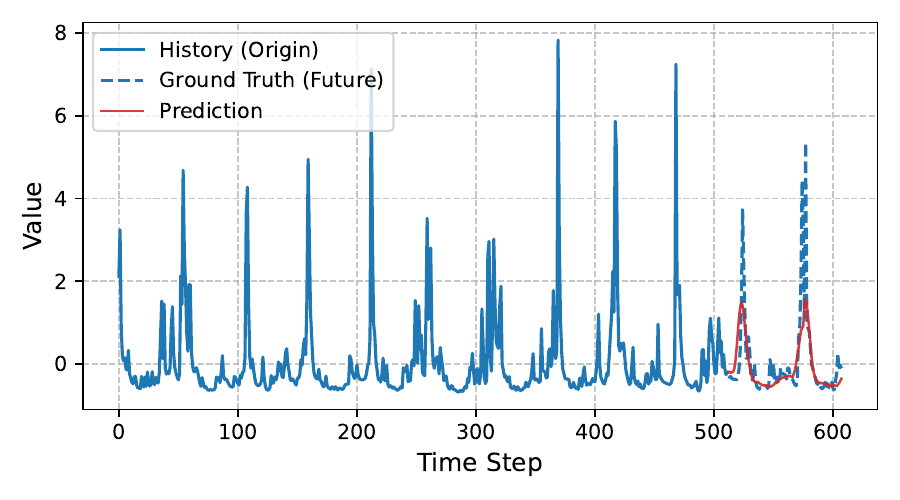}
        \caption{Saugeen-2}
        \label{fig:sub_l}
    \end{subfigure}

    \caption{Example of forecasts from $\method_{b}$ on the test datasets used in experiments.}
    \label{fig:all_subfigures}
\end{figure}

We present several prediction examples generated by \method during testing, as illustrated in Figure \ref{fig:all_subfigures}.

\section{Broader impacts}
\label{app:broader_impacts}
Our work on \method is foundational research focused on advancing parameter-efficient and adaptive time series modeling. We anticipate several positive societal impacts: (i) promoting sustainable AI by significantly reducing the computational resources and energy required for large-scale deployment; (ii) enhancing decision-making in dynamic environments (e.g., energy, retail) through improved modeling of heterogeneous temporal patterns; and (iii) providing a high-quality data curation methodology via the PreSTS corpus to improve model performance.

While we do not foresee immediate negative societal consequences, we recognize that advanced predictive technologies carry inherent risks of misuse. Consequently, rigorous ethical oversight is therefore essential to guarantee that such predictive capabilities serve the public interest.


\end{document}